\documentclass[11pt]{article}

\usepackage[preprint]{acl}

\usepackage{times}
\usepackage{latexsym}
\usepackage[T1]{fontenc}
\usepackage[utf8]{inputenc}
\usepackage{microtype}
\usepackage{inconsolata}
\usepackage{graphicx}

\usepackage{multicol}
\usepackage{algpseudocode} 
\usepackage{tcolorbox}
\tcbuselibrary{skins, breakable, theorems}
\usepackage{amsmath}
\usepackage{amssymb}
\usepackage{amsthm}
\usepackage{booktabs}
\usepackage{framed} 
\usepackage{multirow}
\usepackage{enumitem}
\usepackage{caption}
\usepackage{float} 
\usepackage{etoolbox}
\usepackage{xspace}
\usepackage{accents}
\usepackage{subfig}
\usepackage{colortbl}
\usepackage{xcolor}
\usepackage{hyperref}
\usepackage[ruled,vlined]{algorithm2e} 
\usepackage{array} 
\usepackage{placeins}
\usepackage{marvosym}
\usepackage{tabularx}
\usepackage{makecell}
\usepackage{xurl}
\newtheorem{prob}{Problem}

\newcommand{\eg}{\emph{e.g.},\xspace}

\newcommand{\ie}{\emph{i.e.},\xspace}

\newcommand{\model}{LiPUP-MA}

\newcommand\figref[1]{Figure~\ref{#1}}

\newcommand\tabref[1]{Table~\ref{#1}}
\newcommand\secref[1]{Section~\ref{#1}}
\newcommand\appref[1]{Appendix~\ref{#1}}

\newcommand{\eat}[1]{}

\title{
\model: A Residential Experience-centric Multi-Agent Framework for Living-in-the-loop Participatory Urban Planning}

\author{
 \textbf{Hang Ni\textsuperscript{1}\thanks{Equal contribution.}},
 \textbf{Yuzhi Wang\textsuperscript{2}\footnotemark[1]},
 \textbf{Yizhi Song\textsuperscript{1}},
 \textbf{Hao Liu\textsuperscript{1\ \Letter}
 },
\\
 \textsuperscript{1}The Hong Kong University of Science and Technology (Guangzhou)\\
 \textsuperscript{2}The Hong Kong Polytechnic University
\\
\texttt{hni017@connect.hkust-gz.edu.cn,yu-zhi.wang@connect.polyu.hk}\\
\texttt{ysong531@connect.hkust-gz.edu.cn,liuh@ust.hk}
}

\begin{document}
\maketitle
\begin{abstract}
Participatory Urban Planning (PUP) is increasingly supported by LLM-based agents, yet existing methods largely rely on static preference elicitation and one-shot stakeholder discussions, overlooking the cyclical nature of real-world planning, where residential life, experience collection, and plan adjustment continually interact.
We propose \textbf{Living-in-the-loop Participatory Urban Planning (LiPUP)}, a closed-loop paradigm that alternates between simulated residential living and experience-driven plan revision, while posing two key challenges: grounding scattered living experience in concrete urban contexts and translating subjective feedback into spatially coherent planning actions.
To instantiate LiPUP, we introduce \textbf{\model}, an LLM-based multi-agent framework that constructs a Plan-centric Graph-based Experience Bank to organize urban-grounded residential feedback from living simulation and equips a Spatially-constrained Skill-augmented Planner agent to revise plans by harmonizing experiential, visual, and geospatial evidence.
Experiments show that \model\ consistently outperforms baselines on both conventional static planning metrics and living-based metrics, while iterative LiPUP cycles further improve plan quality.
\end{abstract}

\eat{
Participatory Urban Planning (PUP) is increasingly supported by LLM-based agents, yet existing methods largely rely on static preference elicitation and one-shot stakeholder discussions, overlooking the cyclical nature of real-world planning, where residential life, experience collection, and plan adjustment continually interact.
We propose Living-in-the-loop Participatory Urban Planning (LiPUP), a closed-loop paradigm that alternates between simulated residential living and experience-driven plan revision, while posing two key challenges: grounding scattered living experience in concrete urban contexts and translating subjective feedback into spatially coherent planning actions.
To instantiate LiPUP, we introduce LiPUP-MA, an LLM-based multi-agent framework that constructs a Plan-centric Graph-based Experience Bank to organize urban-grounded residential feedback from living simulation and equips a Spatially-constrained Skill-augmented Planner agent to revise plans by harmonizing experiential, visual, and geospatial evidence.
Experiments show that LiPUP-MA consistently outperforms baselines on both conventional static planning metrics and living-based metrics, while iterative LiPUP cycles further improve plan quality.
}

\section{Introduction}
\label{sec:intro}
Rapid urbanization has rendered many urban configurations increasingly inadequate, making continuous regeneration essential for more livable environments.
Traditional urban planning has largely followed a government-led paradigm, in which planners pursue overarching objectives through regulatory instruments and design guidelines~\cite{taylor1998urban}.
Such top-down approaches, however, often struggle to accommodate heterogeneous land-use preferences among residents.
In response, \textit{Participatory Urban Planning} (PUP) has become central to contemporary planning by involving diverse citizen stakeholders in land-use decision-making, which complements top-down objectives~\cite{ramezani2023comparative}.
Nevertheless, conventional PUP typically relies on surveys, workshops, interviews, and community-specific case studies, making it labor-intensive, difficult to scale, and dependent on domain expertise~\cite{qian2023ai}.
Although Agent-based Modeling (ABM) works have long been used to simulate urban planners and residents~\cite{hosseinali2013agent,qian2023ai}, its agents usually depend on expert-defined rules, which limits adaptive reasoning over resident preferences.
Recent advances in Large Language Models (LLMs) extend this ABM tradition into LLM-based PUP, where language-grounded planner and resident agents can embody diverse profiles, reason over preferences, and support collaborative land-use decisions~\cite{zhou2024large,yong2026intelli}.

\begin{figure}[t]
    \centering
    \includegraphics[width=\linewidth]{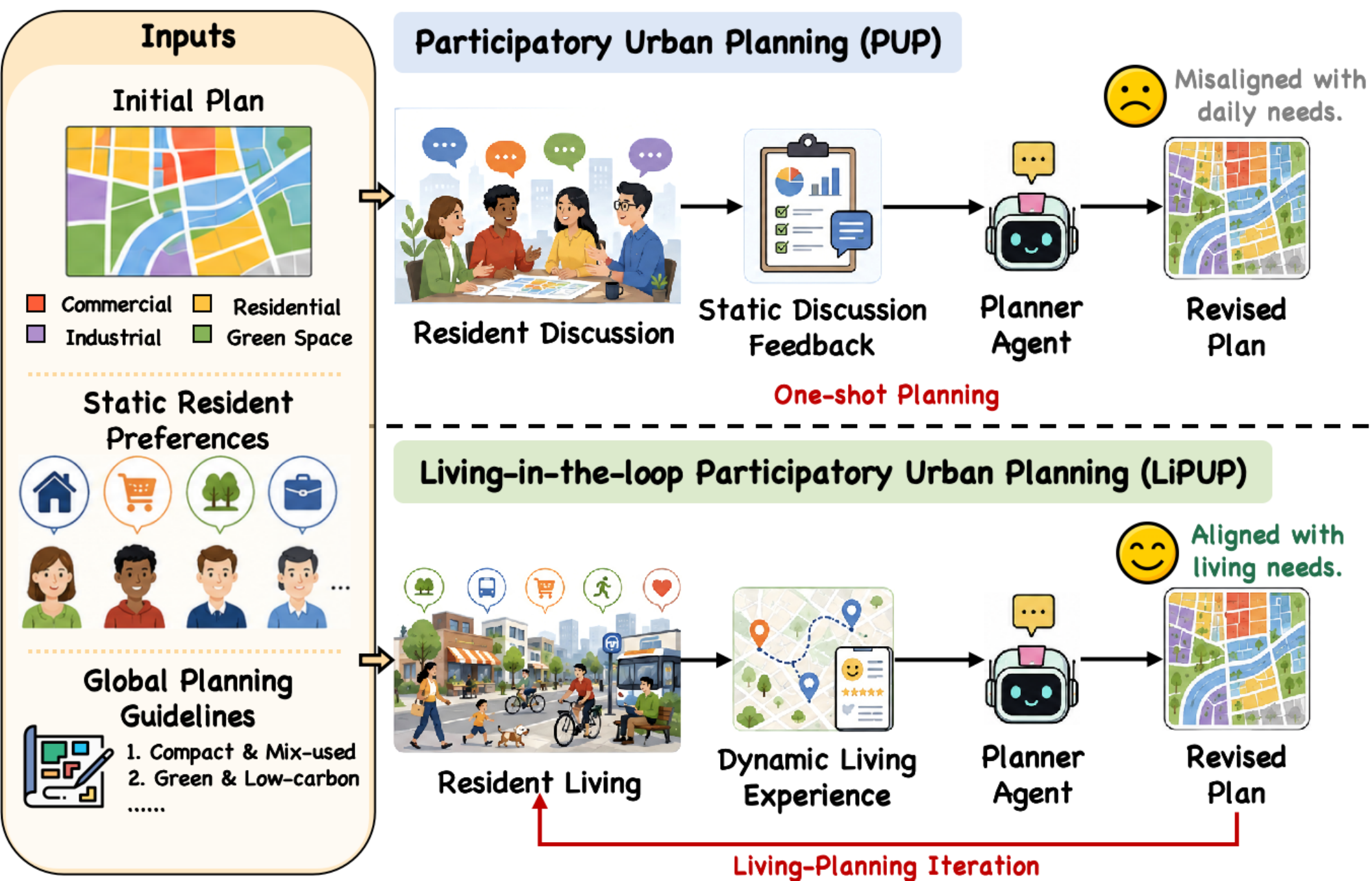}
    \vspace{-20pt}
    \caption{From conventional PUP to LiPUP.}
    \vspace{-20pt}
    \label{fig:cup}
\end{figure}

Despite these advances, real-world PUP is not a one-shot decision-making process.
It instead unfolds through sustained monitoring of residential life, collection of lived experience, and plan adjustment, forming a closed loop that gradually aligns urban configurations with evolving residential needs~\cite{oliveira2010evaluation}.
Existing LLM-based PUP methods, however, often simulate collective decision-making through shallow discussions over static, coarse-grained preferences, without allowing resident agents to inhabit and interact with the urban environment.
Consequently, generated plans may diverge from residents' everyday needs and provide limited experiential evidence for subsequent refinement.
To address this gap, we introduce \textbf{Living-in-the-loop PUP} (LiPUP), a cyclical extension of conventional PUP that incorporates situated residential experience into long-term urban renewal, as shown in \figref{fig:cup}.
Enabled by LLM-based resident and planner agents, LiPUP alternates between two stages:
(1) \textit{Living}, where resident agents simulate daily activities in the planned urban environment and generate plan-grounded living experience; and (2) \textit{Planning}, where planner agents revise the urban plan by integrating global guidelines, static residential preferences, and dynamic living feedback.

However, realizing LiPUP poses two critical challenges:
\textit{(1) How to efficiently ground resident experience in urban contexts?}
A plausible source of living experience is the simulation backbone's built-in resident-wise memory streams.
However, urban living experience emerges from concrete mobility activities through which residents visit and evaluate urban areas~\cite{mouratidis2021urban}, whereas such memories are often dominated by verbose daily trivialities and intermittent generic reflections, making them scattered, redundant, insufficiently tied to concrete urban areas, and even prone to hallucinations.
It is therefore challenging to derive compact, urban-grounded residential experience from mobility activities, which is essential for efficient and reliable plan improvement.
\textit{(2) How to translate living experience into spatially constrained plan revisions?}
Living experience provides bottom-up participatory evidence for the planner agent, but it is inherently localized, subjective, and often biased toward individual needs.
Naively acting on such feedback may violate top-down constraints on the global urban layout or conflict with local spatial feasibility (\eg excessive nearby duplication).
Therefore, the planner must reconcile resident experience with global layout patterns and geospatial details, producing updates that are resident-responsive, globally coherent, and spatially plausible.

To address these challenges, we propose \textbf{\model}, an LLM-empowered multi-agent framework for LiPUP that unifies urban-grounded living simulation and experience-driven plan adaptation within a closed loop.
In the living stage, we propose a \textit{Plan-centric Graph-based Experience Bank} (PGeB) to derive urban-grounded experience from resident simulation.
PGeB elicits spatially bounded reflections from residents' daily activities and organizes them into an experience graph indexed by residents and urban areas, converting fragmented, generic individual memories into structured, area-specific experiential evidence for faithful plan updating.
In the planning stage, we address experience-to-action translation with a \textit{Spatially-constrained Skill-augmented Planner} (S$^2$Planner) agent.
Equipped with multimodal skills, the planner accesses and composes experiential, visual, and geospatial evidence, allowing residential feedback to be interpreted with respect to both the macro functional structure of the urban layout and the micro spatial conditions of candidate areas.
It further regulates plan revision through a locate-ground-execute procedure that progressively translates multimodal evidence into actionable and verified updates.
In addition, we introduce \textit{SimEval}, a simulation-based evaluation protocol that assesses both the objective accessibility of residents' daily mobility and their subjective satisfaction with urban living experiences.

In summary, our contributions are three-fold:
(1) We introduce LiPUP, a new PUP paradigm that closes the loop between living simulation and plan revision, together with the LLM-based framework \model\ and living-based evaluation protocol SimEval.
(2) We propose PGeB and S$^2$Planner to derive urban-grounded residential experience and spatially constrained plan updates, respectively.
(3) Experiments show that \model\ achieves state-of-the-art performance, while iterative LiPUP cycles further enhance plan quality.

\section{Related Work}
\label{sec:related_work}

\noindent\textbf{AI-assisted Urban Planning.}
Recent advances in Artificial Intelligence (AI) have reshaped urban planning as a computational design task, spanning both top-down optimization and participatory paradigms.
Top-down methods optimize urban configurations from a global planning perspective using evolutionary optimization~\cite{koenig2020integrating}, adversarial learning~\cite{wang2023automated}, reinforcement learning~\cite{zheng2023spatial}, or vision-language models~\cite{zhu2025plangpt}.
More recently, participatory planning has emerged as a fundamental principle for urban renewal, shifting attention beyond purely top-down mandates.
Qian et al.~\cite{qian2023ai} develop a consensus-driven multi-agent reinforcement learning (RL) framework to model stakeholder interactions, while LLM-based studies simulate planners and residents for dialogue-driven decision-making~\cite{zhou2024large,zhang2025leveraging}.
Intelli-Planner~\cite{yong2026intelli} further combines LLM reasoning with RL for personalized planning under stakeholder constraints.
However, these methods mostly rely on one-shot discussions over static, coarse-grained preferences, rather than continuously refining plans through simulated lived experiences.
Although Zheng et al.~\cite{zheng2025urban} envision LLM-based living simulations for plan evaluation and refinement, concrete mechanisms for closed-loop adaptation remain underdeveloped.

\noindent\textbf{LLM-based Resident Simulation.}
Recent advances in LLMs have stimulated growing interest in simulating residents within spatial and social environments~\cite{gao2024large}. 
One line of work focuses on mobility-oriented simulation, where LLM agents generate individualized and realistic human movement trajectories~\cite{wang2024large,bhandari2024urban,ju2025trajllm}.
These methods demonstrate the promise of LLMs for modeling fine-grained mobility intentions, but their primary focus remains on movement patterns rather than residents' broader lived experiences.
Another line of work explores more comprehensive society-oriented simulation, where generative agents reproduce human-like daily activities through internal states, long-term memory, and social interaction~\cite{wang2023humanoid,park2023generative,wang2025simulating}.
Recent frameworks further extend this direction to large-scale urban and societal simulations, enabling many LLM-driven agents to jointly model collective behavior and city dynamics~\cite{zhang2025parallelized,bougie2025citysim,ye2026mobilecity}.
Nevertheless, existing methods still offer limited mechanisms for extracting planning-oriented residential experience grounded in concrete spatial units of an urban plan. 
This limitation makes them insufficient for faithful plan assessment and iterative urban renewal.



\section{Problem Statement}
\label{sec:pre}
\begin{prob}
\noindent\textbf{Participatory Urban Planning (PUP).}
Given an urban region $\mathcal{A}$ partitioned into $\{a_1,\dots,a_{N_a}\}$,
an urban plan is defined as a mapping $P: \mathcal{A} \rightarrow \mathcal{U}$,
where each area $a_i$ is assigned a land-use type $u_i = P(a_i)$ and
$\mathcal{U}$ denotes the set of predefined land-use categories.
The goal is to derive an optimized plan $P^\star$ from an initial plan $P_0$ by jointly accounting for top-down planning objectives and residents' static preferences.
\end{prob}


\begin{figure*}[h!]
    \centering
    \includegraphics[width=\linewidth]{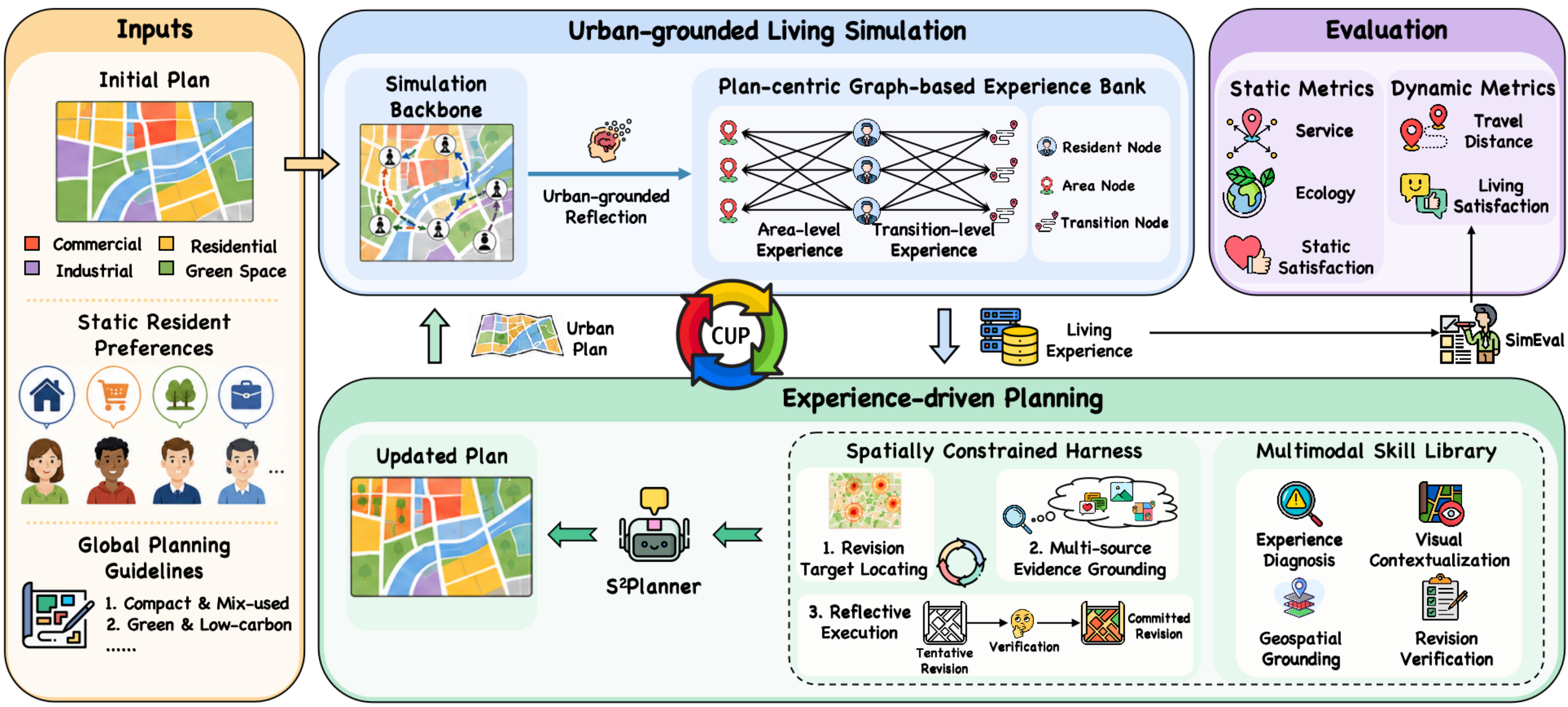}
    \vspace{-20pt}
    \caption{The overall architecture of \model.}
    \vspace{-10pt}
    \label{fig:main}
\end{figure*}

\section{\model}
\label{sec:method}
\figref{fig:main} illustrates the overall architecture of \model, which operates through two alternating stages: \textit{Urban-grounded Living Simulation} and \textit{Experience-driven Planning}.
In the living stage, resident agents simulate daily life under the current urban plan, and their mobility-based reflections are elicited and organized into a graph-structured experience bank to support subsequent plan improvement.
In the planning stage, a \textit{S$^2$Planner} agent revises the urban plan by selectively invoking a multimodal skill library over experiential, visual, and geospatial evidence, while a spatially constrained harness structures its open-ended decision process to iteratively locate revision targets, ground candidate changes, and commit verified land-use updates.
Formally, starting from the initial plan $P_0$ defined in \secref{sec:pre}, LiPUP iterates $E_t=g(P_t)$ and $P_{t+1}=f(P_t,E_t)$ for $t=0,\ldots,T-1$, where $E_t$ is the PGeB-organized living experience produced by simulator $g$ and $f$ denotes S$^2$Planner; after $T$ cycles, the final revised plan is $P^\star=P_T$.

\subsection{Urban-grounded Living Simulation}
\label{sec:living}
To obtain living experience grounded in the current urban plan, we extend resident simulation with a \textit{Plan-centric Graph-based Experience Bank} (PGeB).
Rather than directly using redundant and generic resident-wise memories as planning feedback, PGeB aims to derive constructive experience entries about the efficacy of concrete urban areas from residents' reflections on daily mobility trajectories.
These large-scale entries are then organized into a compact graph structure that explicitly grounds interactions between residents and urban areas, enabling efficient retrieval and aggregation during plan assessment and adaptation.
PGeB can be seamlessly plugged into existing resident simulation frameworks without modifying their underlying simulation dynamics.
In this paper, we instantiate resident simulation on top of AgentSociety~\cite{zhang2025parallelized}, a mature LLM-driven simulation framework.
The following subsections introduce two key designs of PGeB: experience organization and experience storage.

\subsubsection{Experience Organization}
PGeB organizes residential experience as a multiplex graph centered on urban areas, containing three types of nodes:
\textit{(1) Resident nodes} represent simulated residents instantiated with heterogeneous profiles.
\textit{(2) Area nodes} represent the basic urban plan units to be evaluated and potentially revised.
\textit{(3) Transition nodes} represent directed area-to-area movements, each abstracting a relation from an origin area to a destination area.
They bridge residents' movement trajectories and the static urban layout: by modeling transitions as nodes, PGeB goes beyond isolated area-wise remarks and captures how residents experience movement between areas, revealing whether the areas are well connected or lead to inconvenient travel.

PGeB further defines three types of edges to store planning-relevant experience.
\textit{(1) Area-experience edges} connect residents with areas and record area-specific feedback, covering aspects such as convenience, comfort, service adequacy, and conflicts with nearby land uses.
\textit{(2) Transition-experience edges} connect residents with transition nodes and record how a movement is experienced, including its purpose, cost, and perceived friction.
\textit{(3) Spatial-reference edges} connect each transition node with its origin and destination areas, anchoring the movement relation to concrete spatial units.
Through this organization, scattered resident-wise reflections are transformed into plan-indexed experience records that support efficient aggregation by area, resident group, and inter-area transition.

\subsubsection{Experience Storage}
To populate PGeB with useful experience, we introduce an urban-grounded reflection behavior that is triggered whenever a resident completes a movement from area $a_i$ to area $a_j$.
Instead of producing a generic diary-style reflection on mundane events, the resident is prompted to draw experience from the completed mobility event with respect to the graph entities it involves: the origin area $a_i$, the destination area $a_j$, and the transition $a_i \rightarrow a_j$.
The generated feedback entries are conditioned on the resident's profile, historical experience, travel intention, and observed environmental context, which concern either the areas themselves or their surrounding land-use configurations.
For the $q$-th mobility record $v_{m,q}$ of resident $r_m$ under plan $P_t$, whose movement goes from $a_i$ to $a_j$, the urban-grounded reflections elicit two experience types:
\[
\begin{aligned}
e^{A}_{m,q,k}&=\rho_A(r_m,a_k,v_{m,q},P_t),\quad k\in\{i,j\},\\
e^{T}_{m,q,i,j}&=\rho_T(r_m,a_i\rightarrow a_j,v_{m,q},P_t),
\end{aligned}
\]
where $e^{A}_{m,q,k}$ is area-level experience for area $a_k$, $e^{T}_{m,q,i,j}$ is transition-level experience from $a_i$ to $a_j$, and $\rho_A,\rho_T$ denote the corresponding reflection functions.
They are then written to the corresponding area-experience and transition-experience edges in PGeB, converting daily mobility into faithful, spatially grounded evidence for later plan assessment and refinement.

\subsection{Experience-driven Planning}
\label{sec:planning}
To translate living experience into globally coherent and spatially feasible land-use updates, the planning stage centers on a \textit{Spatially-constrained Skill-augmented Planner} (S$^2$Planner), a skill-based autonomous agent equipped with a \textit{Multimodal Skill Library} (MSL) and regulated by a \textit{Spatially Constrained Harness} (SCH).
MSL addresses the evidence side of the experience translation challenge by packaging experiential, visual, geospatial, and verification capabilities as reusable multimodal skills, enabling the planner to interpret bottom-up residential feedback together with macro-level layout patterns and micro-level spatial facts.
SCH addresses the decision-process side of this challenge by constraining open-ended agent reasoning into iterative target selection, evidence collection, and trial-based execution steps, preventing the planner from directly turning biased and localized complaints into edits.

\subsubsection{Multimodal Skill Library}
MSL provides four skills, each defined by an \textit{operating protocol} and a \textit{tool set}: the protocol specifies how the capability should be used, while the tool set provides executable operations.
\textit{(1) Experience Diagnosis} treats PGeB as bottom-up planning evidence: it retrieves hotspot areas, area-level dissatisfaction, neighbor externalities, and transition frictions, then distills them into recurring resident needs and the planning issues behind them.
\textit{(2) Visual Contextualization} supplies the macro-level insight missing from textual complaints by rendering global and local maps, highlighting target areas, and captioning visible patterns, so the planner can relate local problems to functional clusters, service gaps, and layout coherence.
\textit{(3) Geospatial Grounding} turns visual impressions into precise spatial constraints by querying area attributes, adjacency, distances, service coverage, and land-use compatibility, thereby separating feasible intervention space from redundant or risky changes.
\textit{(4) Revision Verification} closes the evidence-action loop by applying tentative edits, checking constraints and cumulative changes, reading metric feedback, and rolling back harmful trials before a revision is accepted.
Together, these skills let the planner diagnose resident needs, understand their spatial context, delimit feasible interventions, and verify final plan updates.
The details of these skills are provided in \appref{app:skill}.

\subsubsection{Spatially Constrained Harness}
SCH structures S$^2$Planner into step-wise skill-using loops, where each loop contains three steps with distinct objectives, available skills, and expected intermediate outputs, while the planner flexibly decides which skill to invoke and when the step is complete.
This design regulates the transition from feedback to action while preserving adaptive skill use within each spatially constrained step.

\noindent\textbf{Locate-Ground-Execute Loop.}
SCH organizes planning into repeated locate, ground, and execute steps.
In \textit{(1) Revision Target Locating}, the planner maintains a global target list and selects priority targets by combining resident experience with visual layout inspection, where each target is a small cluster of neighboring areas.
In \textit{(2) Multimodal Evidence Grounding}, it collects and organizes experiential, visual, and geospatial evidence around selected targets, clarifying resident needs and multi-view spatial constraints.
Visual and geospatial information combine macro-level layout perception with micro-level spatial details: the former reveals how candidate areas relate to surrounding functional patterns, while the latter provides neighborhood context and distance-based spatial cues.
In \textit{(3) Reflective Execution}, it uses grounded evidence to repeatedly propose tentative edits and verify them through temporary updates and metric feedback.
The planner can accept beneficial changes or roll back harmful ones until the revision budget and planning objectives are satisfied.
This loop requires the planner to produce explicit intermediate decisions: where to revise, why a revision is spatially supported, and whether the resulting update should be committed.
Formally, within planning cycle $t$, SCH starts from $P_t^{(0)}=P_t$ and runs loop $\ell$ as $z_\ell=\operatorname{Locate}(P_t^{(\ell)},E_t)$, $h_\ell=\operatorname{Ground}(P_t^{(\ell)},E_t,z_\ell)$, and $P_t^{(\ell+1)}=\operatorname{Execute}(P_t^{(\ell)},z_\ell,h_\ell)$, where $z_\ell$ is the selected revision target set and $h_\ell$ is the evidence collected for it; after $L$ loops, $P_{t+1}=P_t^{(L)}$.

\noindent\textbf{Skill-Tool Hierarchical Reasoning.}
Within each step, the planner separates decision making into two nested levels, composing high-level skills across the step while keeping low-level tool calls governed by skill-specific protocols rather than a flat, unconstrained tool list.
At the \textit{skill-selection} level, the planner observes the step-wise objective, skill-level history, and available skills, then decides either to invoke the next skill from MSL or to finish the current step with the required output.
Once a skill is selected, the planner enters the \textit{skill-execution} level, where it follows the selected skill's operating protocol and iteratively invokes the corresponding tools until a compact skill result is produced.
Formally, the planner selects $s=\pi_{\mathrm{skill}}(c,\mathcal{S})$ from the skill set $\mathcal{S}$ under context $c$, executes tools as $(o_1,\ldots,o_K)=\pi_{\mathrm{tool}}(c,s,\Omega_s)$ from the tool set $\Omega_s$, and returns $y=\sigma_s(o_{1:K})$, where $o_k$ are tool observations and $\sigma_s$ summarizes them into the compact skill result.

\noindent\textbf{Context Management.}
To support long-horizon revision, SCH maintains two levels of context aligned with the reasoning hierarchy. 
The \textit{loop-level context} tracks global revision progress across loops, storing completed step results from previous loops and the skills already executed within the current step.
When the context approaches the length limit, earlier loop trajectories are compressed into compact summaries.
In contrast, the \textit{skill-local context} records fine-grained execution details within the currently selected skill, including tool calls and their observations.
After the skill finishes, its detailed trajectory is summarized into a skill result and returned to the skill-level context.
This separation preserves global planning state across loops while preventing verbose tool traces from overwhelming subsequent decisions.

\section{SimEval}
\label{sec:simeval}

Beyond top-down planning constraints and residents' static land-use preferences, we propose \textit{SimEval} to evaluate the living-oriented quality of an urban plan under the LiPUP paradigm. 
SimEval measures how well a plan supports residents' simulated daily lives from two complementary perspectives: objective mobility efficiency and subjective experiential satisfaction.

\definecolor{initgray}{HTML}{F0F0F0}
\definecolor{baseblue}{HTML}{E8ECF8}
\definecolor{oursred}{HTML}{FBEAEA}
\definecolor{bestblue}{HTML}{003399}
\definecolor{bestred}{HTML}{B00000}

\begin{table*}[!t]
\centering
\caption{Performance comparison results across two urban regions.
\textcolor{bestblue}{\textbf{Blue bold}} indicates the best-performing baseline method, while \textcolor{bestred}{\textbf{red bold}} indicates the best-performing cycle of \model.}
\vspace{-10pt}
\resizebox{\linewidth}{!}{
\scriptsize
\begin{tabular}{c|c|c c c|c|c c}
\toprule
\multicolumn{1}{c|}{\multirow{2}{*}{\textbf{Region}}}
& \multicolumn{1}{c|}{\multirow{2}{*}{\textbf{Method}}}
& \multicolumn{4}{c|}{\textbf{Static Metrics}}
& \multicolumn{2}{c}{\textbf{Dynamic Metrics}} \\
\cmidrule{3-6}\cmidrule{7-8}
\multicolumn{1}{c|}{}
& \multicolumn{1}{c|}{}
& \textbf{Service$\ \uparrow$}
& \textbf{Ecology$\ \uparrow$}
& \textbf{Static Satis.$\ \uparrow$}
& \textbf{Overall$\ \uparrow$}
& \textbf{Travel Dist.$\ \downarrow$}
& \textbf{Living Satis.$\ \uparrow$} \\

\midrule

\multirow{8}{*}{\textbf{Beijing}}
& \cellcolor{initgray}\textbf{Initial}
& \cellcolor{initgray}0.4106 & \cellcolor{initgray}0.4317 & \cellcolor{initgray}0.3684 & \cellcolor{initgray}0.4036
& \cellcolor{initgray}447.92 & \cellcolor{initgray}2.5722 \\

\cmidrule{2-8}

& \cellcolor{baseblue}\textbf{Random}
& \cellcolor{baseblue}\textcolor{bestblue}{\textbf{0.5339}} 
& \cellcolor{baseblue}0.3150 
& \cellcolor{baseblue}0.4302 
& \cellcolor{baseblue}0.4264
& \cellcolor{baseblue}426.96 
& \cellcolor{baseblue}2.5445 \\

& \cellcolor{baseblue}\textbf{GA}
& \cellcolor{baseblue}0.5228 
& \cellcolor{baseblue}0.5550 
& \cellcolor{baseblue}0.4858 
& \cellcolor{baseblue}0.5212
& \cellcolor{baseblue}\textcolor{bestblue}{\textbf{399.96}} 
& \cellcolor{baseblue}2.4944 \\

& \cellcolor{baseblue}\textbf{DRL-GNN}
& \cellcolor{baseblue}0.5128 
& \cellcolor{baseblue}\textcolor{bestblue}{\textbf{0.6200}} 
& \cellcolor{baseblue}\textcolor{bestblue}{\textbf{0.5287}} 
& \cellcolor{baseblue}\textcolor{bestblue}{\textbf{0.5538}}
& \cellcolor{baseblue}423.47 
& \cellcolor{baseblue}2.5967 \\

& \cellcolor{baseblue}\textbf{PUP-MA}
& \cellcolor{baseblue}0.4394 
& \cellcolor{baseblue}0.6017 
& \cellcolor{baseblue}0.5237 
& \cellcolor{baseblue}0.5216
& \cellcolor{baseblue}407.97 
& \cellcolor{baseblue}\textcolor{bestblue}{\textbf{2.6056}} \\

\cmidrule{2-8}

& \cellcolor{oursred}\textbf{\model$_{\ \text{cycle 1}}$}
& \cellcolor{oursred}0.5289 
& \cellcolor{oursred}\textcolor{bestred}{\textbf{0.7083}} 
& \cellcolor{oursred}0.5979 
& \cellcolor{oursred}0.6117
& \cellcolor{oursred}370.94 
& \cellcolor{oursred}2.6078 \\

& \cellcolor{oursred}\textbf{\model$_{\ \text{cycle 2}}$}
& \cellcolor{oursred}0.7278 
& \cellcolor{oursred}\textcolor{bestred}{\textbf{0.7083}} 
& \cellcolor{oursred}0.7800 
& \cellcolor{oursred}0.7387
& \cellcolor{oursred}366.60 
& \cellcolor{oursred}2.6578 \\

& \cellcolor{oursred}\textbf{\model$_{\ \text{cycle 3}}$}
& \cellcolor{oursred}\textcolor{bestred}{\textbf{0.7694}} 
& \cellcolor{oursred}\textcolor{bestred}{\textbf{0.7083}} 
& \cellcolor{oursred}\textcolor{bestred}{\textbf{0.8046}} 
& \cellcolor{oursred}\textcolor{bestred}{\textbf{0.7608}}
& \cellcolor{oursred}\textcolor{bestred}{\textbf{364.09}} 
& \cellcolor{oursred}\textcolor{bestred}{\textbf{2.6867}} \\

\midrule

\multirow{8}{*}{\textbf{Berlin}}
& \cellcolor{initgray}\textbf{Initial}
& \cellcolor{initgray}0.3589 & \cellcolor{initgray}0.5367 & \cellcolor{initgray}0.2736 & \cellcolor{initgray}0.3897
& \cellcolor{initgray}651.79 & \cellcolor{initgray}2.5733 \\

\cmidrule{2-8}

& \cellcolor{baseblue}\textbf{Random}
& \cellcolor{baseblue}0.4900 & \cellcolor{baseblue}0.5367 & \cellcolor{baseblue}0.3453 & \cellcolor{baseblue}0.4573
& \cellcolor{baseblue}654.74 & \cellcolor{baseblue}2.5833 \\

& \cellcolor{baseblue}\textbf{GA}
& \cellcolor{baseblue}\textcolor{bestblue}{\textbf{0.5117}} & \cellcolor{baseblue}0.5367 & \cellcolor{baseblue}0.3667 & \cellcolor{baseblue}0.4717
& \cellcolor{baseblue}\textcolor{bestblue}{\textbf{614.94}} & \cellcolor{baseblue}2.5425 \\

& \cellcolor{baseblue}\textbf{DRL-GNN}
& \cellcolor{baseblue}0.4744 & \cellcolor{baseblue}0.5667 & \cellcolor{baseblue}\textcolor{bestblue}{\textbf{0.3939}} & \cellcolor{baseblue}\textcolor{bestblue}{\textbf{0.4783}}
& \cellcolor{baseblue}656.00 & \cellcolor{baseblue}2.8933 \\

& \cellcolor{baseblue}\textbf{PUP-MA}
& \cellcolor{baseblue}0.4900 & \cellcolor{baseblue}\textcolor{bestblue}{\textbf{0.5883}} & \cellcolor{baseblue}0.3389 & \cellcolor{baseblue}0.4724
& \cellcolor{baseblue}640.68 & \cellcolor{baseblue}\textcolor{bestblue}{\textbf{2.9100}}\\

\cmidrule{2-8}

& \cellcolor{oursred}\textbf{\model$_{\ \text{cycle 1}}$}
& \cellcolor{oursred}0.5922 & \cellcolor{oursred}0.5367 & \cellcolor{oursred}0.4200 & \cellcolor{oursred}0.5163
& \cellcolor{oursred}622.56 & \cellcolor{oursred}3.0167 \\

& \cellcolor{oursred}\textbf{\model$_{\ \text{cycle 2}}$}
& \cellcolor{oursred}0.6061 & \cellcolor{oursred}\textcolor{bestred}{\textbf{0.8017}} & \cellcolor{oursred}0.5117 & \cellcolor{oursred}0.6398
& \cellcolor{oursred}570.72 & \cellcolor{oursred}3.1008 \\

& \cellcolor{oursred}\textbf{\model$_{\ \text{cycle 3}}$}
& \cellcolor{oursred}\textcolor{bestred}{\textbf{0.6372}} & \cellcolor{oursred}\textcolor{bestred}{\textbf{0.8017}} & \cellcolor{oursred}\textcolor{bestred}{\textbf{0.5975}} & \cellcolor{oursred}\textcolor{bestred}{\textbf{0.6788}}
& \cellcolor{oursred}\textcolor{bestred}{\textbf{563.59}} & \cellcolor{oursred}\textcolor{bestred}{\textbf{3.1225}} \\

\bottomrule
\end{tabular}
}
\vspace{-10pt}
\label{tab:comparison}
\end{table*}

\noindent\textbf{Travel Distance.}
This metric quantitatively evaluates residents' daily mobility burden by computing the average travel distance incurred by all resident agents during living simulation. 
Compared with static area-accessibility measures based only on land-use distribution, it further incorporates fine-grained activity intentions and realized movement trajectories, thereby providing a more behavior-grounded estimate of urban convenience.

\noindent\textbf{Living Satisfaction.}
This metric qualitatively assesses residents' subjective approval of the urban plan by querying each resident agent with an LLM-as-a-Judge protocol based on its simulated daily experiences~\cite{park2024generative}. 
Each resident assigns an integer satisfaction score from 1 to 5, and we average the ratings across all residents to capture collective perceptions shaped by mobility activities and social interactions, rather than merely reflecting static land-use preference matching.
Metric details are provided in \appref{app:metric}, and the human evaluation results in \appref{app:humaneval} further demonstrate that the LLM-based evaluators align well with human judgments.

\section{Experiment}
\label{sec:exp}

\begin{figure*}[h!]
    \centering
    \includegraphics[width=\linewidth]{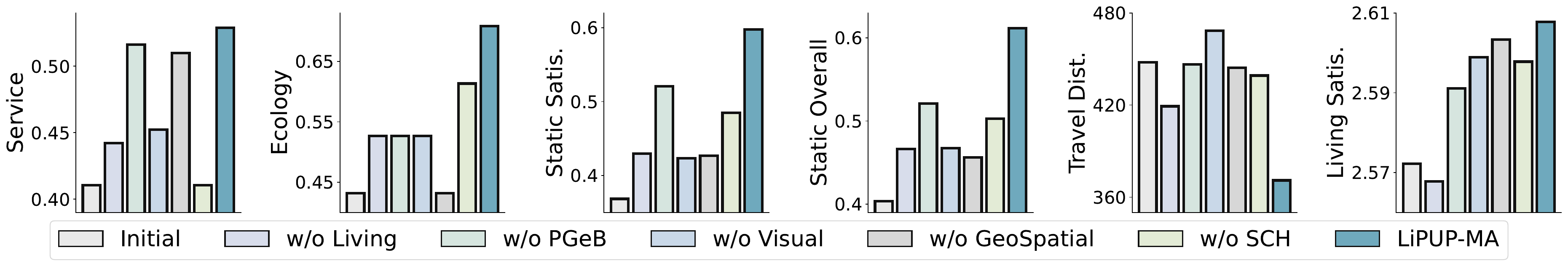}
    \vspace{-20pt}
    \caption{Ablation results of \model\ in Daxing, Beijing.}
    \vspace{-10pt}
    \label{fig:ablation}
\end{figure*}

\subsection{Experimental Setup}

\noindent\textbf{Study Regions.}
We conduct experiments on community-level regions in two cities:
\textit{(1) Beijing}, the capital of China, where we select a region in the Daxing Biomedical Industry Base. The region spans [116$^\circ$17$'$46$''$E, 39$^\circ$41$'$25$''$N] to [116$^\circ$19$'$12$''$E, 39$^\circ$40$'$09$''$N], covers approximately 2.02\,km $\times$ 2.35\,km, and consists of 68 areas, of which 55 are modifiable.
\textit{(2) Berlin}, the capital of Germany, where we select a region in Adlershof. The region spans [13$^\circ$31$'$45$''$E, 52$^\circ$26$'$30$''$N] to [13$^\circ$33$'$18$''$E, 52$^\circ$25$'$34$''$N], covers approximately 1.75\,km $\times$ 1.72\,km, and consists of 48 areas, of which 30 are modifiable.

\noindent\textbf{Baselines.}
We compare \model\ with five baselines:
\textit{(1) Initial}, which directly uses the original plan as a reference for measuring absolute improvement;
\textit{(2) Random}, which randomly samples feasible modifications and represents unguided exploration;
\textit{(3) GA}~\cite{koenig2020integrating}, a population-based evolutionary search method that iteratively selects and mutates candidate solutions;
\textit{(4) DRL-GNN}~\cite{zheng2023spatial}, which combines deep reinforcement learning with graph neural networks to capture spatial dependencies and learn planning policies;
and \textit{(5) PUP-MA}~\cite{zhou2024large}, an LLM-based multi-agent planning framework that facilitates planner-resident collaboration using static residential preferences to support balanced planning decisions.

\noindent\textbf{Evaluation Metrics.}
We evaluate urban plan quality using both conventional static metrics and simulation-based dynamic metrics.
The static metrics include:
\textit{(1) Service}~\cite{zheng2023spatial}, which measures the spatial efficiency of public-service layouts by assessing residents' access to essential area types within a neighborhood-scale distance;
\textit{(2) Ecology}~\cite{zheng2023spatial}, which reflects the coverage of green space by measuring whether residents are within the service range of nearby park areas;
and \textit{(3) Static Satisfaction}~\cite{zhou2024large}, which measures the consistency between an urban plan and residents' stated land-use preferences by checking whether preferred area types are conveniently accessible.
The dynamic metrics include \textit{(4) Travel Distance} and \textit{(5) Living Satisfaction}, both derived from our SimEval protocol, as detailed in \secref{sec:simeval}.
Additional metric details are provided in \appref{app:metric}.

In this paper, we iterate our \model\ over three cycles.
Additional implementation details are provided in \appref{app:implement}.

\subsection{Comparison Results}
\tabref{tab:comparison} presents the performance comparison across different urban regions, from which we draw the following key observations:
(1) \model\ consistently outperforms all baselines on both static and dynamic metrics, and already achieves superior performance in the first cycle, demonstrating the effectiveness of explicitly generating and leveraging living experience to improve plan quality.
(2) \model\ exhibits a clear iterative improvement trend across the three cycles, indicating that the proposed LiPUP paradigm can progressively refine urban plans through experience-driven feedback.
(3) ``Random'' generally yields the weakest performance among the baselines, suggesting that feasible urban-plan modification remains a highly non-trivial search problem where unguided exploration is insufficient.
(4) Among the baseline methods, DRL-GNN achieves the strongest performance on most static metrics, reflecting the advantage of reinforcement learning for structural plan optimization, while PUP-MA obtains the best baseline living satisfaction, highlighting the benefit of incorporating participatory preference modeling for improving residents' quality of life.
(5) The inconsistency between the Service metric and Travel Distance results suggests that conventional static accessibility indicators may not fully capture the nuanced mobility costs experienced by residents in dynamic urban environments.
The planning results for each cycle of our \model\ are provided in \appref{app:result}, while a case study is presented in \appref{app:case}.

\subsection{Ablation Study}
We further validate the contribution of each component in \model\ by comparing it with several variants:
``w/o Living'' removes the living stage;
``w/o PGeB'' replaces our PGeB with a naive memory stream, where we apply an LLM-based compression post-processing technique to summarize the experience streams of all residents into compact textual memories;
``w/o Visual'' and ``w/o Geospatial'' remove the visual contextualization and geospatial grounding skills, respectively;
and ``w/o SCH'' removes the locate-ground-execute harness.
\figref{fig:ablation} reports the ablation results, demonstrating:
(1) The living stage is crucial for LiPUP, as removing resident simulation leads to substantially weaker performance, especially on the two dynamic metrics.
(2) Urban-grounded experience representation is essential, as the clear gap between ``w/o PGeB'' and \model\ shows that trivial and redundant memories are insufficient for preserving actionable planning guidance.
(3) Both visual and geospatial skills contribute meaningfully to plan optimization, and the full model achieves the best overall performance by effectively integrating these complementary modalities.
(4) The locate-ground-execute harness plays a critical role in translating experience into executable planning actions, as removing it consistently degrades performance across most metrics.

\subsection{In-depth Experience Analysis}
We further analyze how resident experience contributes to experience-guided cyclical planning in the proposed \model.

\begin{figure}[h!]
    \centering
    \vspace{-5pt}
    \includegraphics[width=0.95\linewidth]{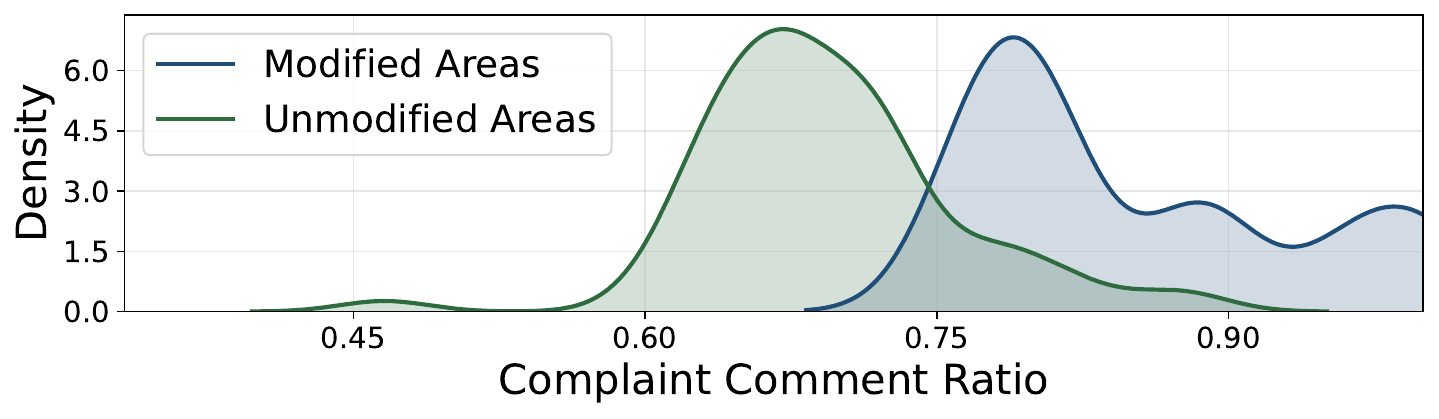}
    \vspace{-10pt}
    \caption{Complaint-comment ratio distributions of modified and unmodified areas.}
    \vspace{-10pt}
    \label{fig:analysis_utilize}
\end{figure}

\noindent\textbf{Experience Utilization Analysis.}
To examine whether \model\ truly leverages the experience bank, we compare the complaint-comment ratio distributions of areas modified during planning with those left unchanged.
As shown in \figref{fig:analysis_utilize}, modified areas generally receive stronger negative feedback, indicating that \model\ incorporates residents' experiential signals.
Meanwhile, the partial overlap suggests that it does not blindly follow complaints, but also accounts for spatial and feasibility constraints.

\begin{figure}[h!]
    \centering
    \vspace{-5pt}
    \includegraphics[width=0.85\linewidth]{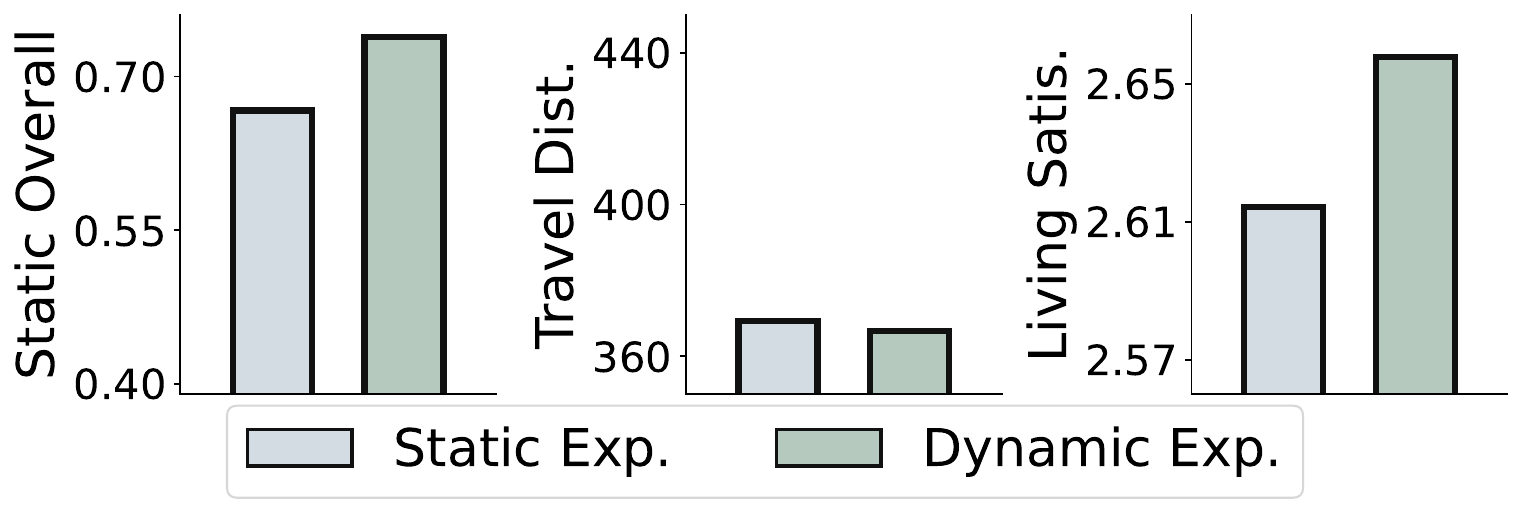}
    \vspace{-10pt}
    \caption{Experience-static Intervention.}
    \vspace{-10pt}
    \label{fig:analysis_static}
\end{figure}

\noindent\textbf{Experience-static Intervention Analysis.}
To verify the necessity of dynamically updated experience, we re-run \model$_{\ \text{cycle 2}}$ by replacing the latest experience bank with the historical experience bank generated from the initial plan.
As shown in \figref{fig:analysis_static}, using static historical feedback leads to inferior performance compared with the standard \model$_{\ \text{cycle 2}}$, demonstrating that experience should be refreshed as the plan evolves to provide effective guidance.

\begin{figure}[h!]
    \centering
    \vspace{-5pt}
    \includegraphics[width=0.85\linewidth]{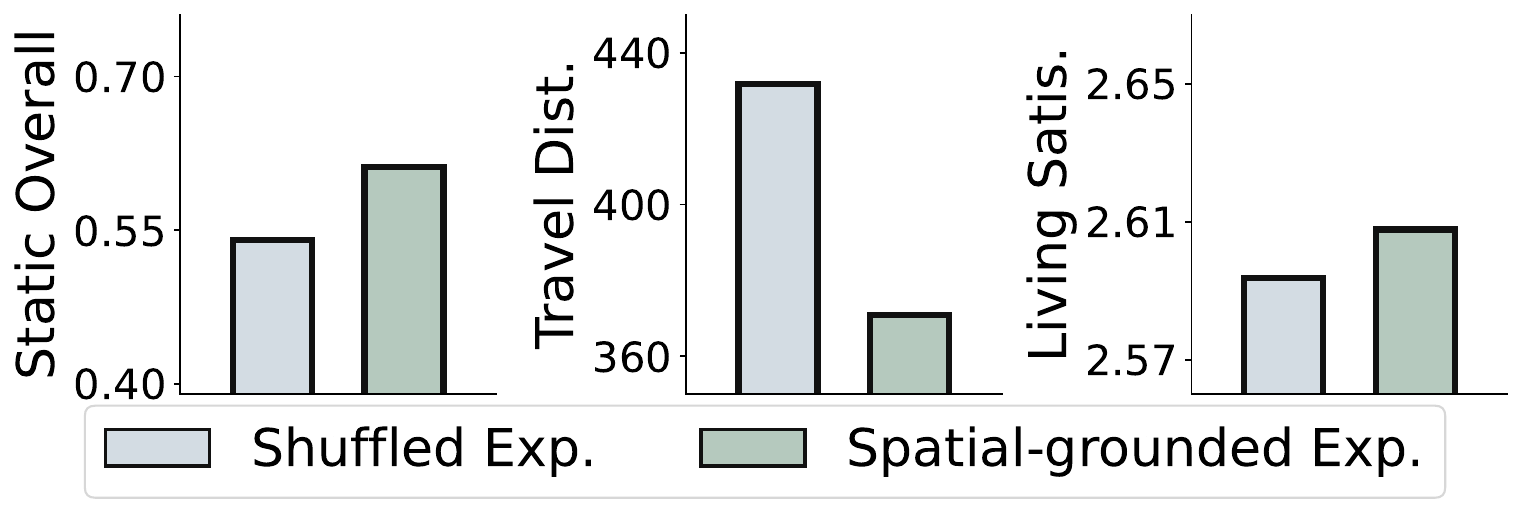}
    \vspace{-10pt}
    \caption{Experience-shuffled Intervention.}
    \vspace{-10pt}
    \label{fig:analysis_shuffled}
\end{figure}

\noindent\textbf{Experience-shuffled Intervention Analysis.}
To assess whether experience must be spatially grounded, we randomly reassign each reflection to a different area while preserving its textual content, and then re-run \model$_{\ \text{cycle 1}}$.
As shown in \figref{fig:analysis_shuffled}, the degradation caused by shuffled experience shows that experiential feedback is effective only when correctly aligned with the urban locations it describes.

\begin{figure}[h!]
    \centering
    \vspace{-5pt}
    \includegraphics[width=0.85\linewidth]{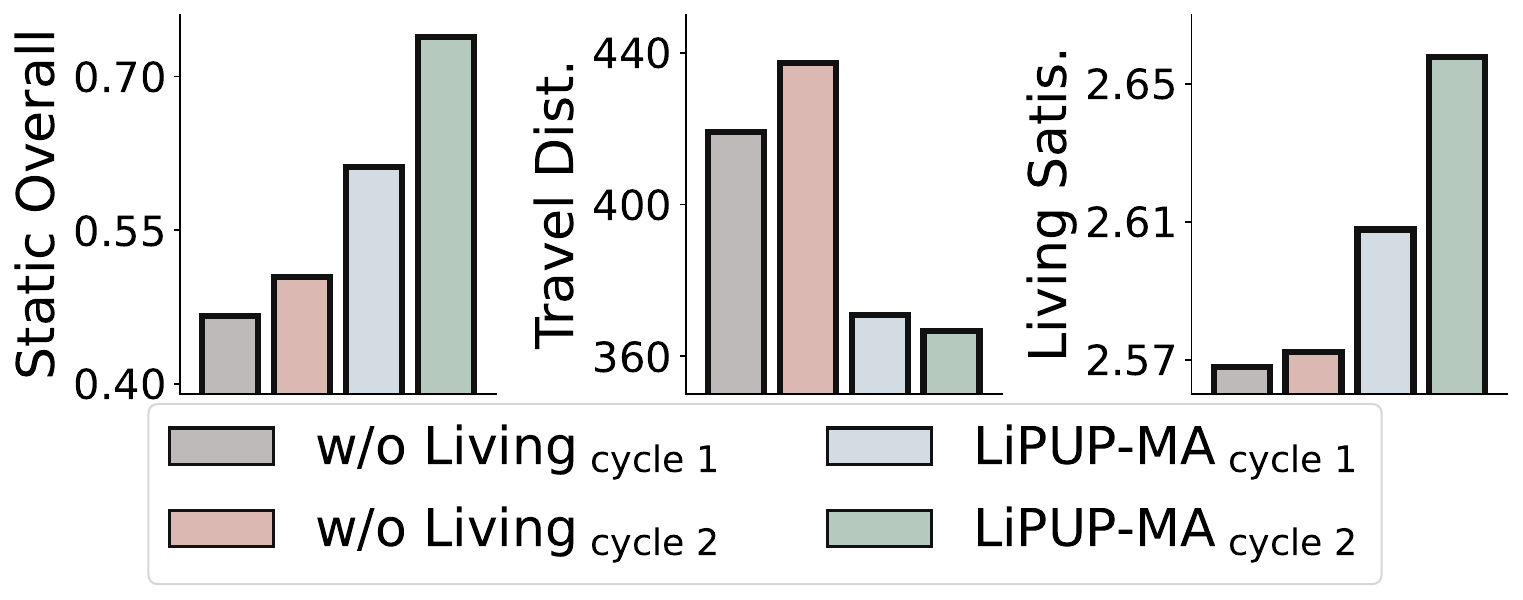}
    \vspace{-10pt}
    \caption{Experience-disabled Cyclical Intervention.}
    \vspace{-10pt}
    \label{fig:analysis_no_live}
\end{figure}

\noindent\textbf{Experience-disabled Cyclical Intervention Analysis.}
To examine whether repeated planning alone can explain the gains of \model, we continue planning from the ``w/o Living'' variant while keeping the living stage disabled.
As shown in \figref{fig:analysis_no_live}, the additional cycle without living experience achieves substantially weaker results than the full \model, suggesting that the benefits of LiPUP largely come from experiential feedback rather than iteration alone.

\section{Conclusion}
We introduced LiPUP, a new LLM-based planning paradigm that closes the loop between residential living simulation and adaptive urban revision, moving beyond static preference-based participatory planning.
\model\ instantiates this paradigm with PGeB, which organizes fragmented resident memories into urban-grounded experiential evidence, and S$^2$Planner, which translates such evidence into spatially constrained plan updates.
Experiments show consistent gains over baselines on both static and simulation-based metrics, and the ablation results confirm the value of grounded experience, multimodal evidence, and constrained revision.
The progressive gains across LiPUP cycles further demonstrate the promise of iterative, living-experience-driven urban adaptation.

\section*{Limitations}

\noindent\textbf{Simulation Accuracy.}
LiPUP uses LLM-based resident simulation as its computational substrate, so its effectiveness in real-world applications depends on how accurately the simulation backbone models residents' lives.
However, improving simulation fidelity is orthogonal to our main research goal, which is to formulate living-in-the-loop PUP as a new optimizable problem and instantiate it with a computationally practical simulation loop.
As urban simulators become more reliable, the proposed framework can be more readily adapted to real-world planning scenarios.

\noindent\textbf{Scope of Urban Planning.}
This work focuses on land-use planning, while urban planning also involves transportation, building density, infrastructure, zoning, and environmental design.
We also keep area geometries fixed and do not edit GeoJSON boundaries, leaving geometry-aware planning to future work.

\noindent\textbf{Regional Scale.}
Our experiments are conducted on two community-level regions for controlled evaluation.
Scaling LiPUP to larger cities, more regions, and richer populations would require substantially higher simulation costs.

\noindent\textbf{Iteration Budget.}
We run three LiPUP cycles because each cycle requires a full living simulation and planning stage.
More cycles may further improve plan quality, and efficient reuse or adaptive stopping remains future work.

\noindent\textbf{Training-Free Planner.}
Our planner is a skill-augmented LLM agent without task-specific agent training.
Reinforcement learning could further improve planning efficiency and long-horizon optimization, which we leave for future work.

\noindent\textbf{Spatiotemporal Generalization.}
This work focuses on iteratively optimizing one planning region.
Future studies can extend LiPUP to long-term planning under evolving resident populations and coordinated planning across multiple regions.

\noindent\textbf{Evaluation Robustness.}
Resident simulation and living-satisfaction evaluation depend on specific LLM backbones.
Future work can test additional model families and calibrate simulated feedback against real survey evidence.

\section*{Ethical Considerations}

This work uses publicly available OpenStreetMap data and aggregate census statistics for the selected regions, and does not involve private records or personally identifiable information. Resident profiles are automatically sampled and synthesized from regional aggregate census distributions, while living experiences and satisfaction judgments are simulation artifacts. The human evaluation only asks annotators to assess simulated outputs without exposing sensitive personal data. For real-world urban governance, \model\ serves as a decision-support aid, and any practical deployment must incorporate professional planners, public deliberation, local regulations, and validation by affected communities. If released, processed data and code will preserve the required attribution and usage conditions of their original public sources.


\bibliography{custom}

\appendix
\begin{figure*}[h!]
    \centering
    \subfloat[Initial.]{\includegraphics[width=0.499\linewidth]{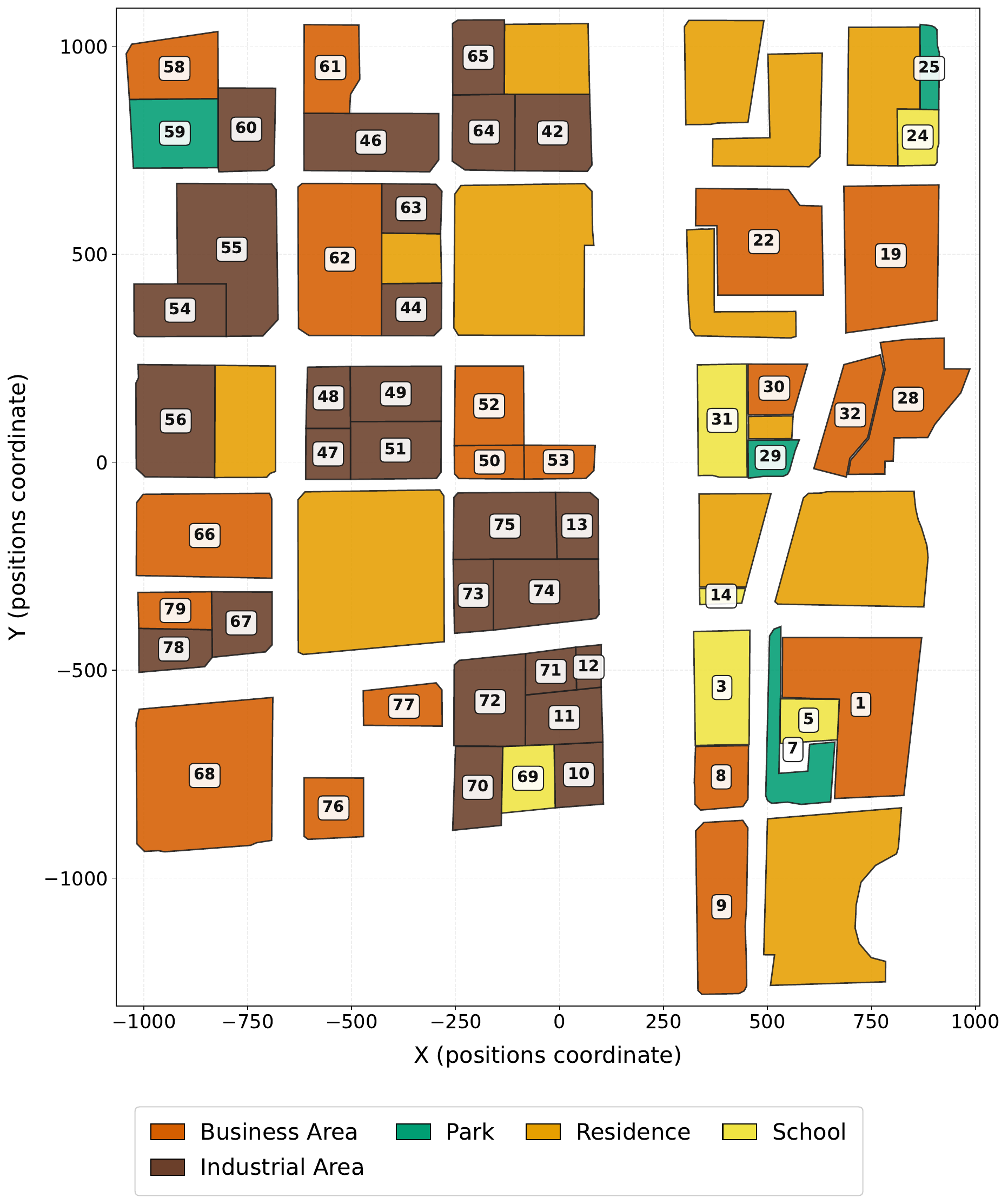}\label{fig:result_bj_0}}
    \subfloat[Cycle 1.]{\includegraphics[width=0.499\linewidth]{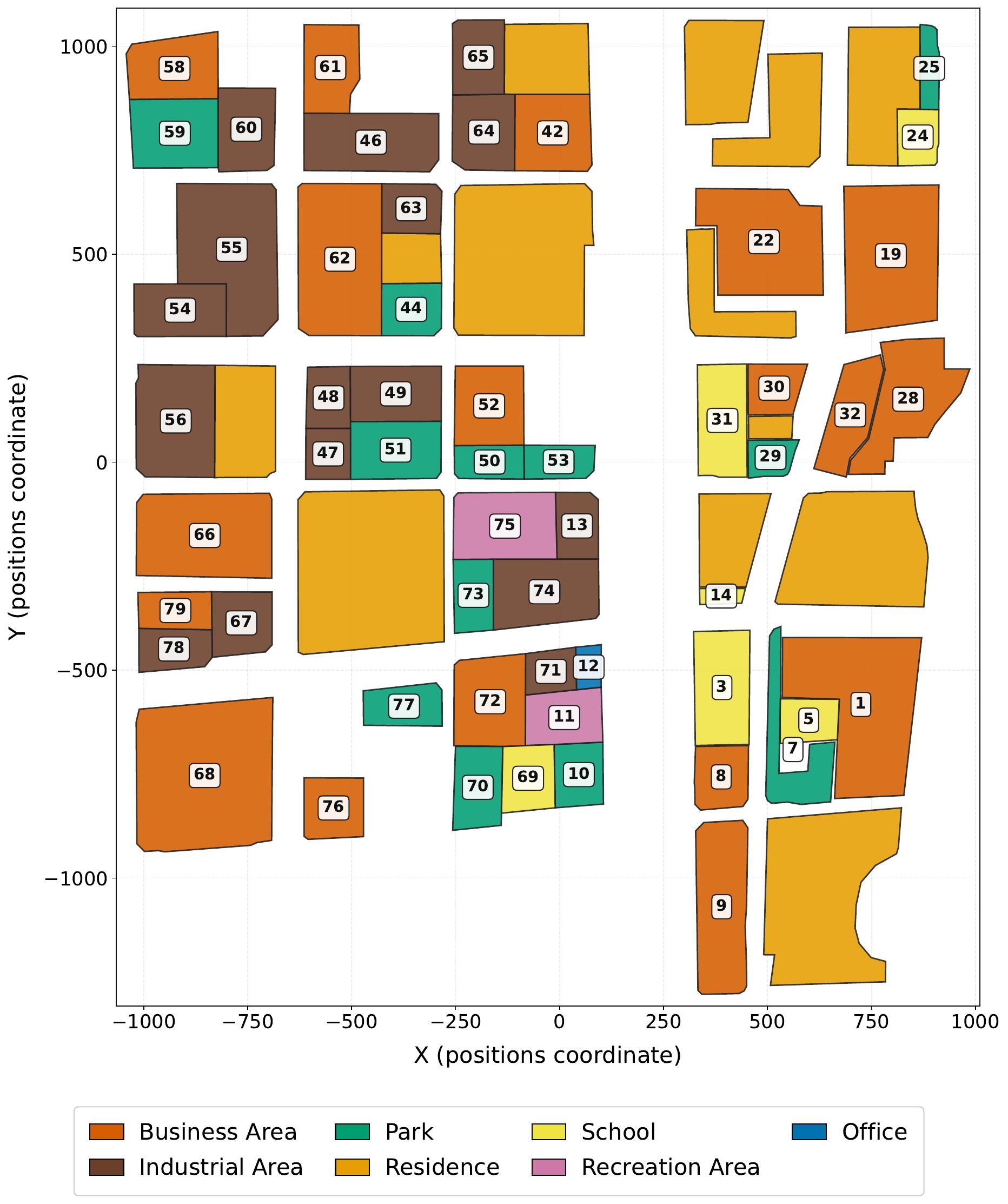}\label{fig:result_bj_1}}

    \subfloat[Cycle 2.]{\includegraphics[width=0.499\linewidth]{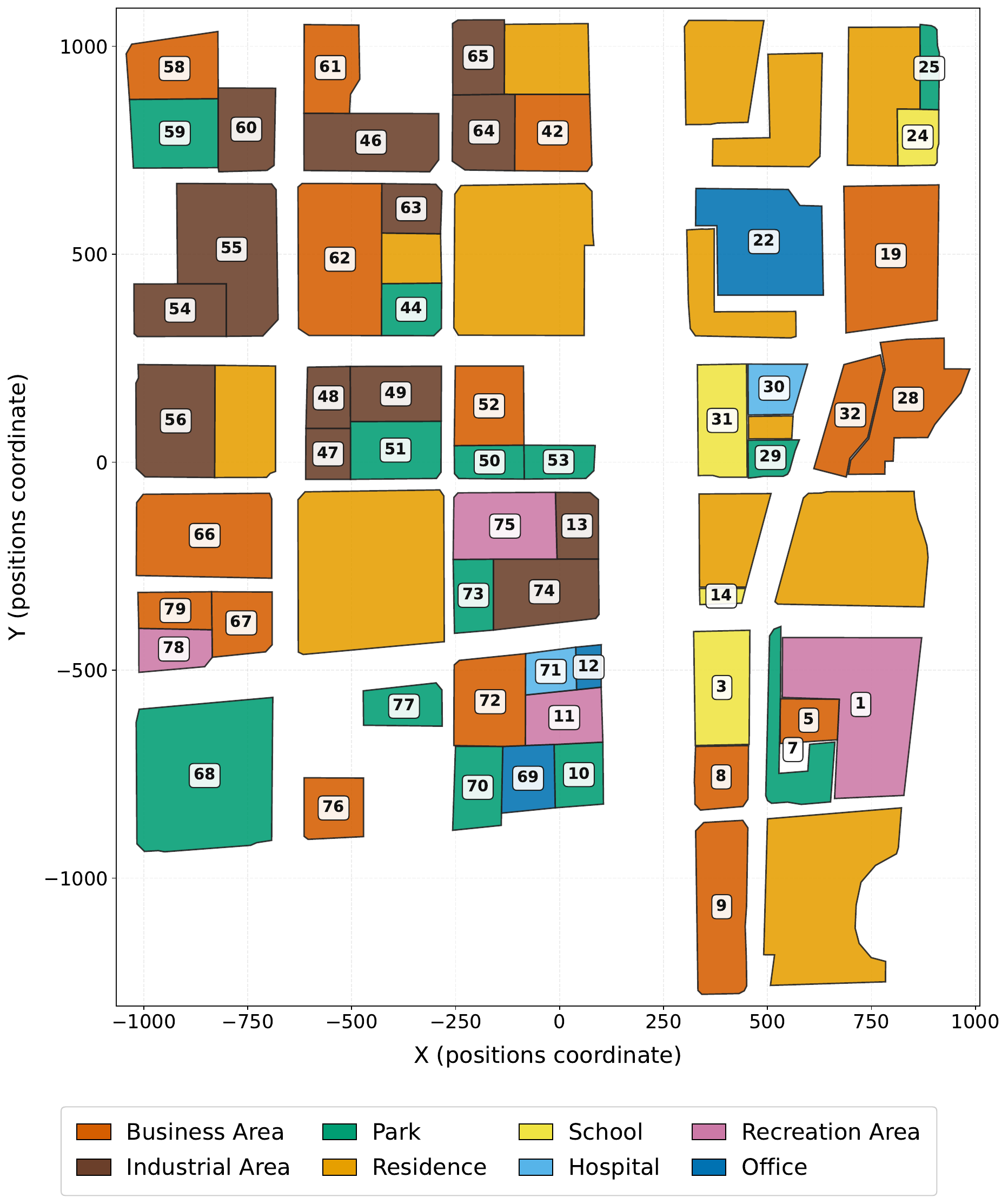}\label{fig:result_bj_2}}
    \subfloat[Cycle 3.]{\includegraphics[width=0.499\linewidth]{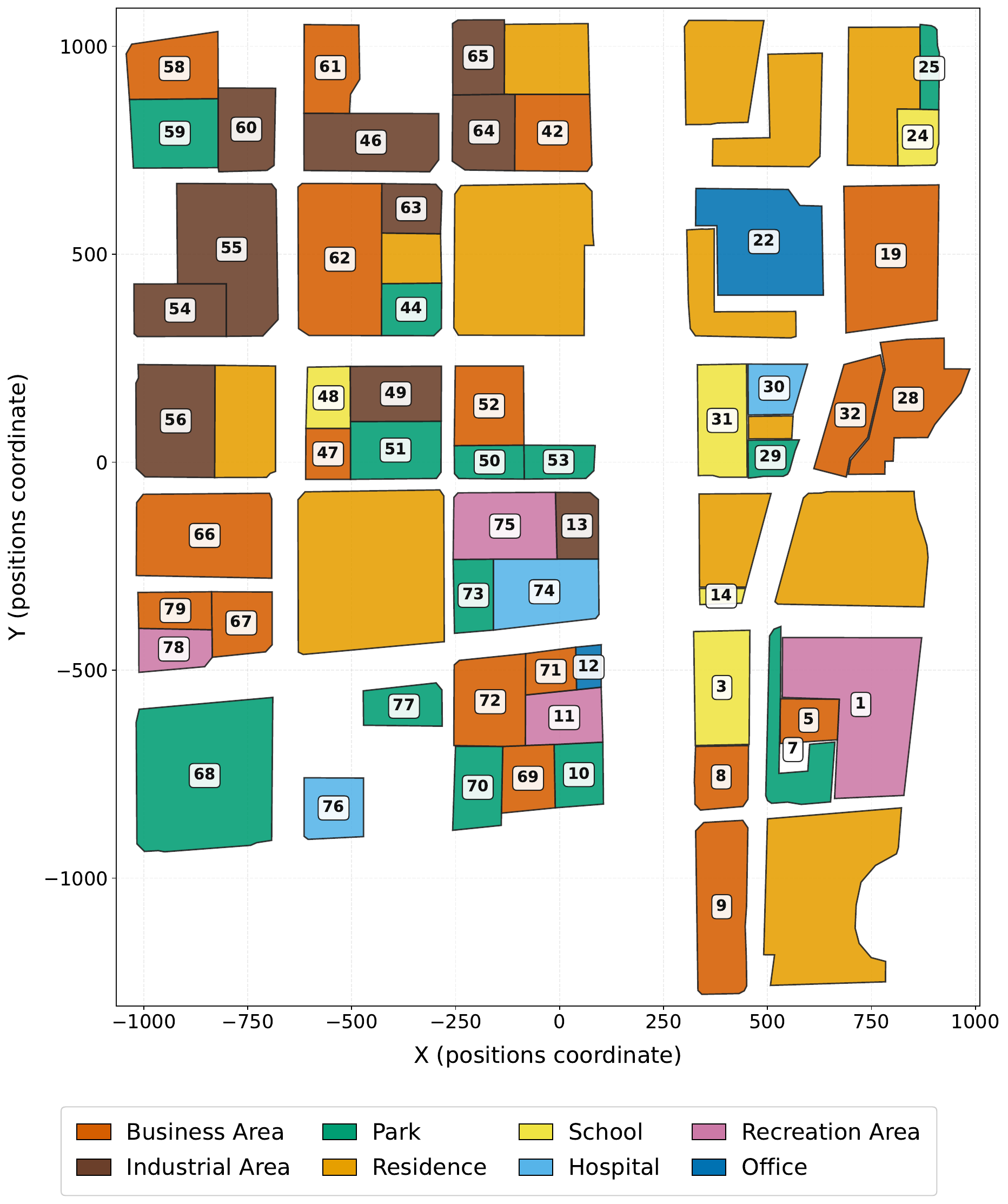}\label{fig:result_bj_3}}
    \caption{Planning results in the Beijing region.}
    \label{fig:result_bj}
\end{figure*}

\begin{figure*}[h!]
    \centering
    \subfloat[Initial.]{\includegraphics[width=0.499\linewidth]{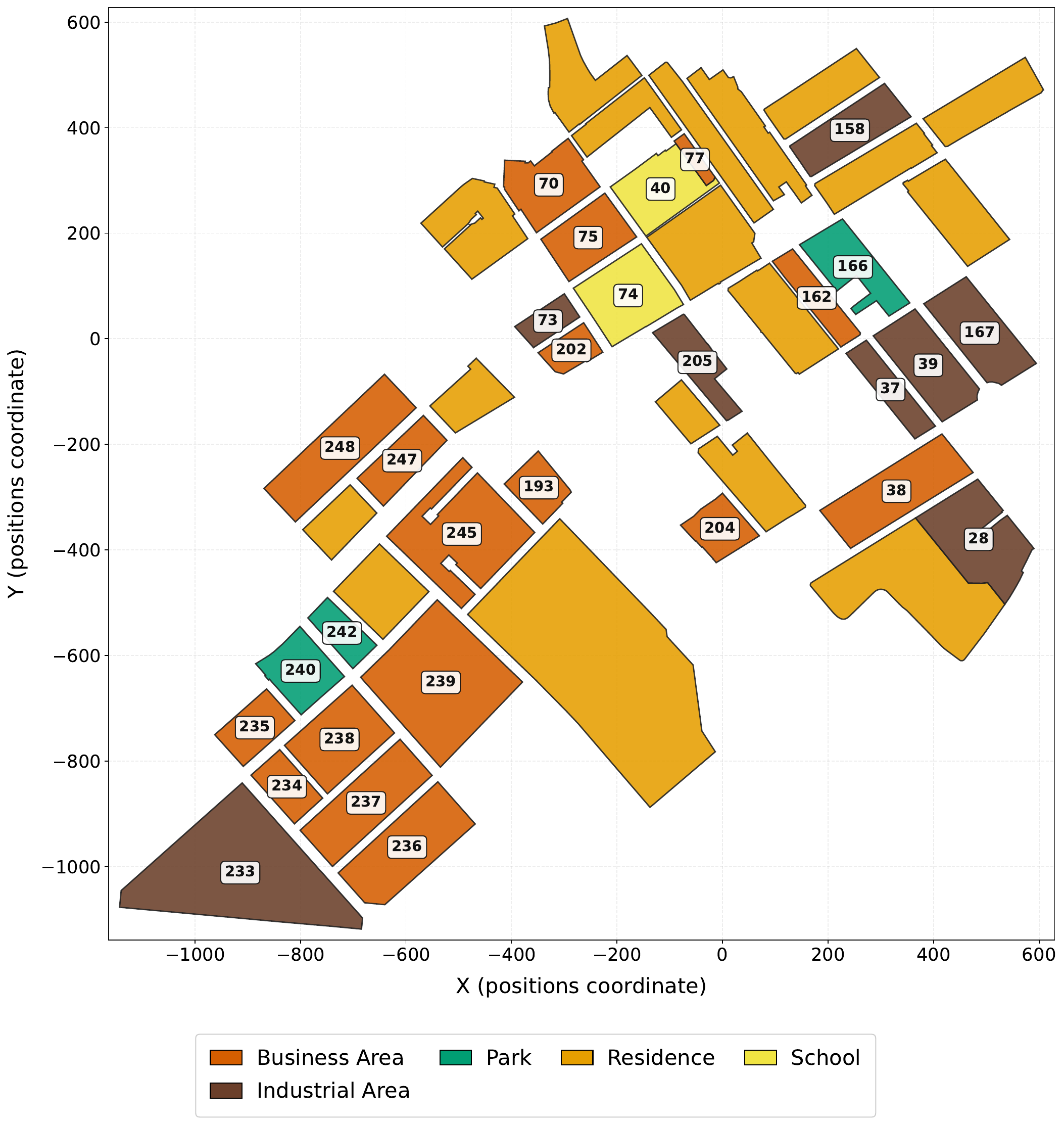}\label{fig:result_bl_0}}
    \subfloat[Cycle 1.]{\includegraphics[width=0.499\linewidth]{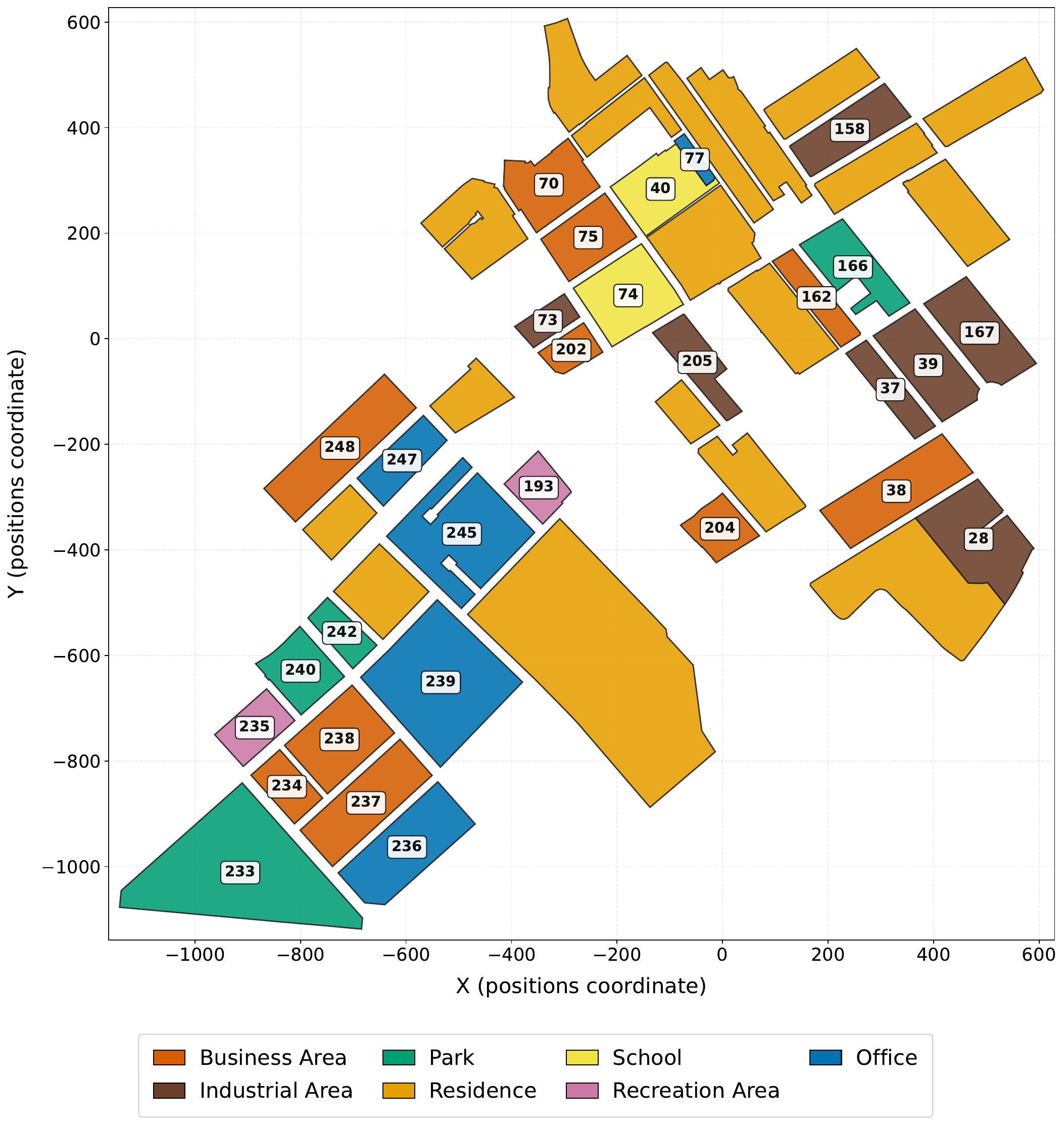}\label{fig:result_bl_1}}

    \subfloat[Cycle 2.]{\includegraphics[width=0.499\linewidth]{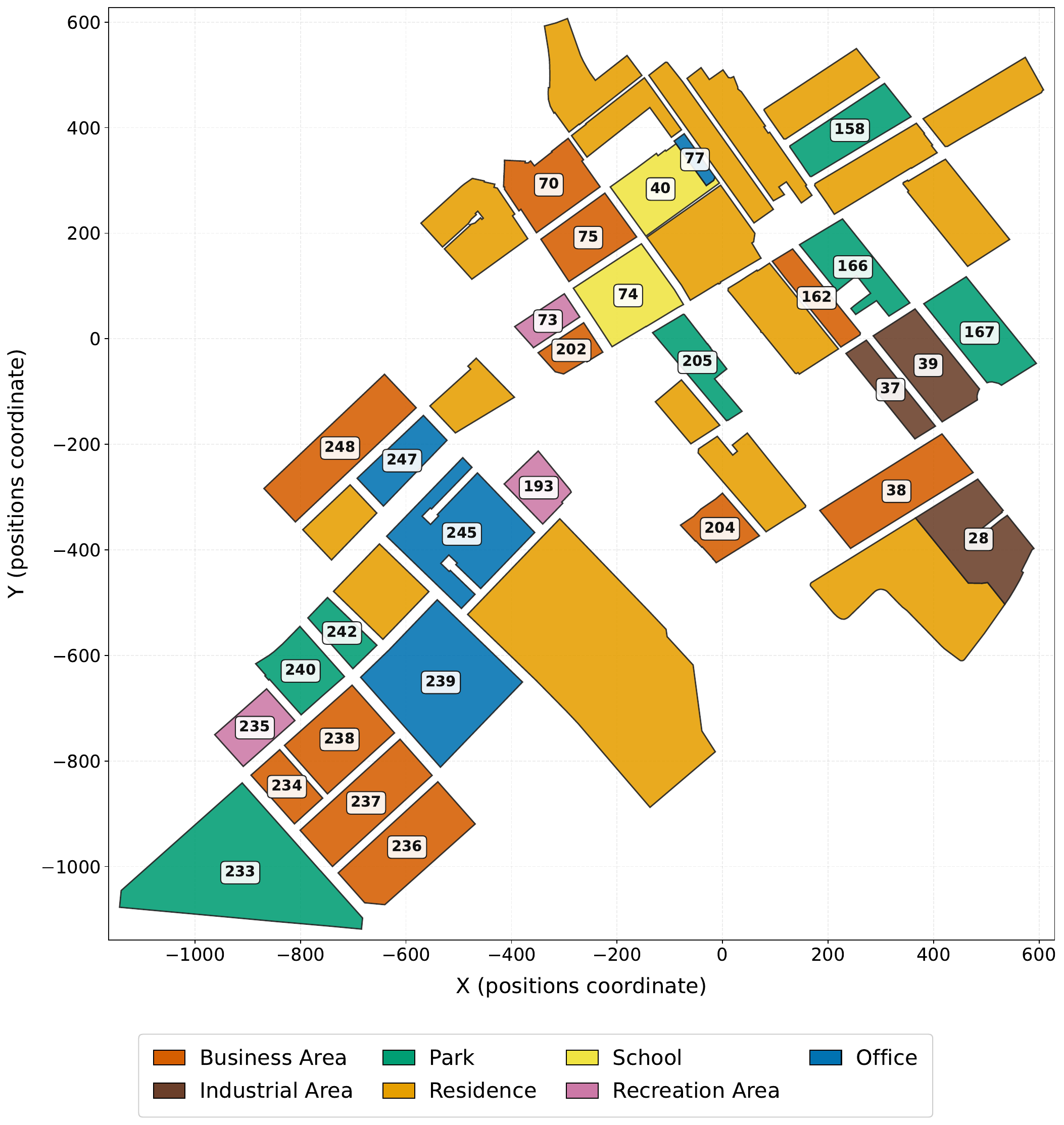}\label{fig:result_bl_2}}
    \subfloat[Cycle 3.]{\includegraphics[width=0.499\linewidth]{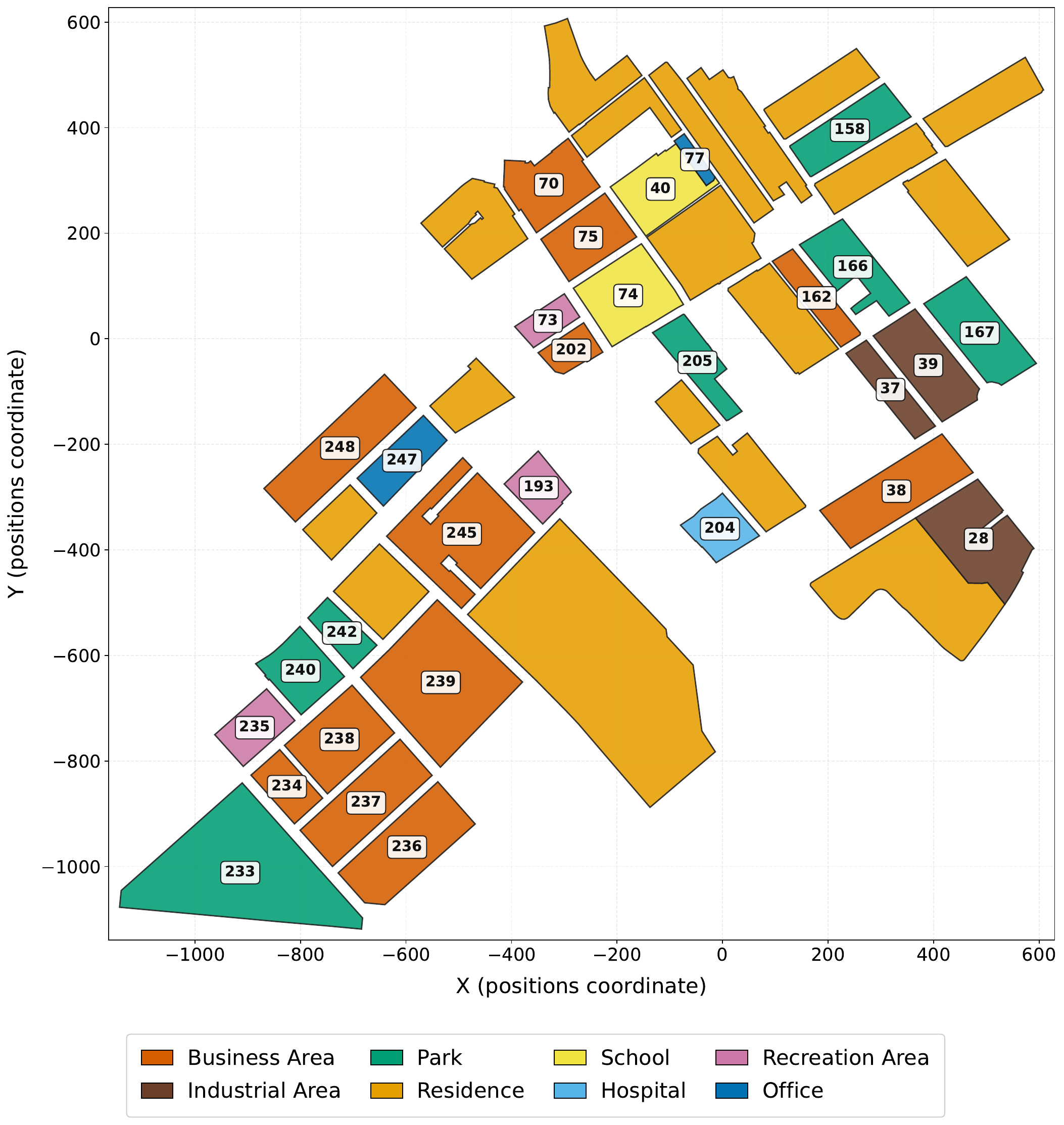}\label{fig:result_bl_3}}
    \caption{Planning results in the Berlin region.}
    \label{fig:result_bl}
\end{figure*}

\begin{figure*}[h!]
    \centering
    \subfloat[Initial.]{\includegraphics[width=0.499\linewidth]{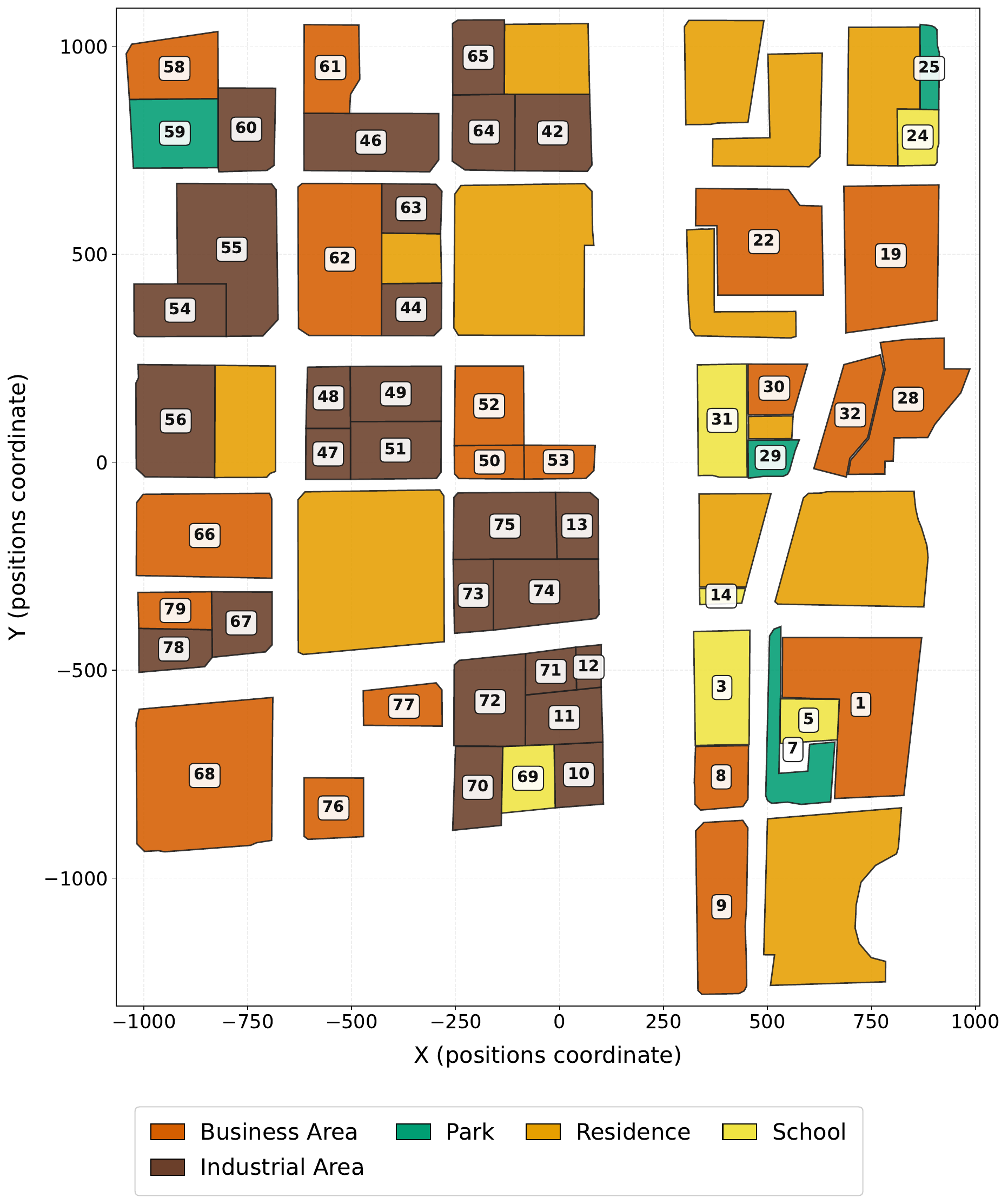}\label{fig:case_0}}
    \subfloat[After Loop 1.]{\includegraphics[width=0.499\linewidth]{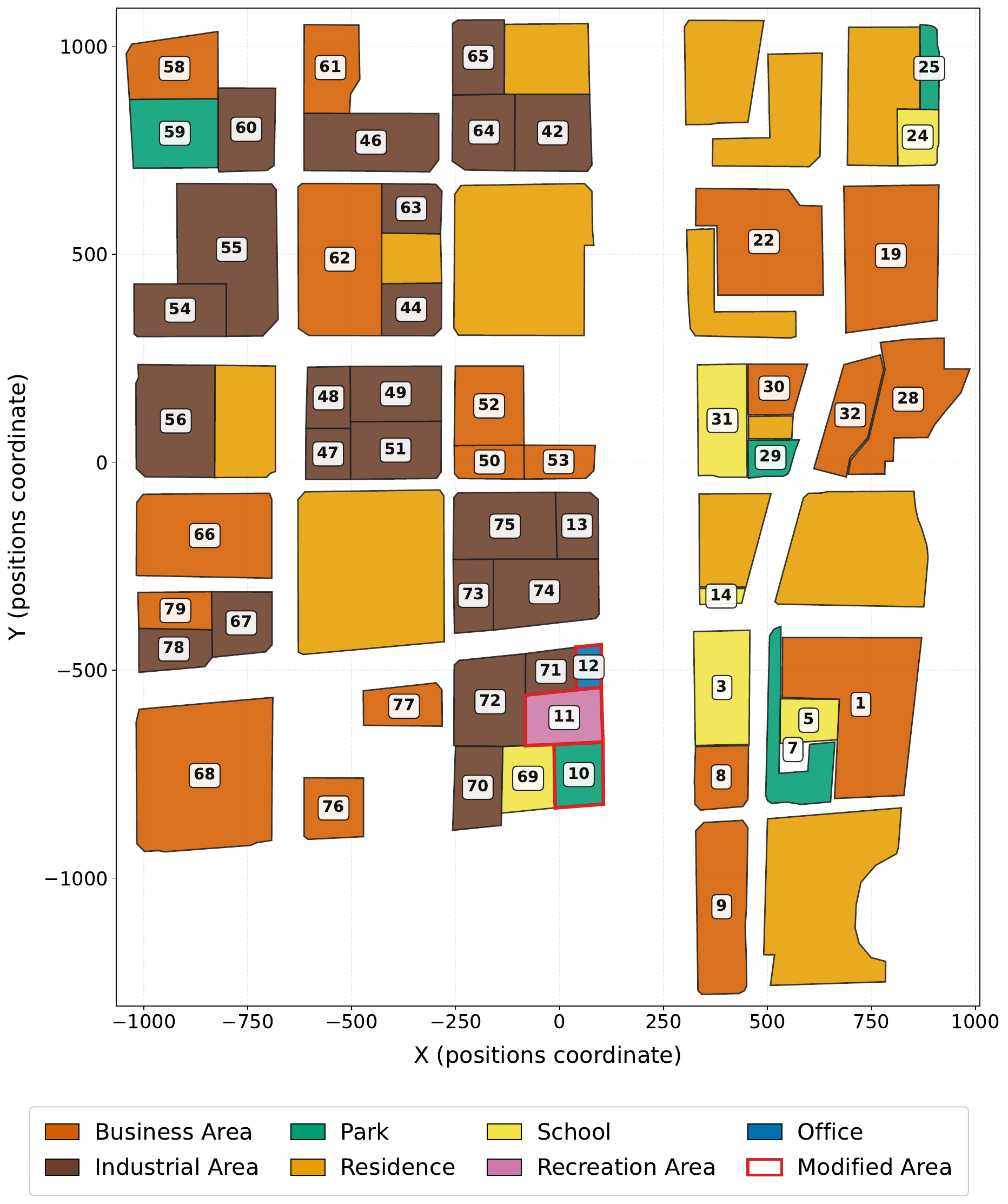}\label{fig:case_1}}
    \caption{Case analysis in the Daxing region. Highlighted AOIs show revisions from industrial use to other land-use types.}
    \label{fig:case}
\end{figure*}

\section{Planning Results}
\label{app:result}

The generated urban plans across three cycles of \model\ are shown in \figref{fig:result_bj} (Beijing) and \figref{fig:result_bl} (Berlin).

\section{Case Study}
\label{app:case}

\figref{fig:case} shows a representative revision made in the Daxing region during the first harness loop of \model$_{\ \text{Cycle 1}}$.
Resident reflections around the highlighted industrial corridor did not simply reject the existing workplace function; instead, workers and nearby residents described it as accessible and useful, yet isolated from everyday amenities, short-break destinations, shaded walking environments, and small social spaces.
The planner then grounded these comments in the surrounding area: nearby schools, residences, parks, and business parcels already provided important local functions, while industrial parcels 10, 11, and 12 formed a concentrated opportunity to add missing neighborhood support without disrupting stronger adjacent uses.
As a result, \model\ converted 10 from industrial use to Park (PK), 11 to Recreation Area (RC), and 12 to Office (OF).
This case illustrates that the framework does not modify areas solely according to complaint frequency; it links resident experience to specific locations, checks neighboring land-use context, and produces a spatially coherent revision that preserves employment activity while adding green, recreation, and service-supportive functions.

\section{Implementation Details}
\label{app:implement}
We provide additional implementation details for the 
experiments. 
We construct the study-region map from OpenStreetMap\footnote{\url{https://www.openstreetmap.org}} data. The raw OSM elements are processed with OSMnx (v1.1.2+) and represented as GeoPandas (v0.14.4+) GeoDataFrames. We standardize land-use annotations by manually mapping inconsistent OSM tags into the predefined land-use categories used in our experiments, merge or remove highly fragmented parcels, and filter out spatial elements that are irrelevant to land-use planning, such as non-planning auxiliary geometries. Geometric operations are implemented with Shapely (v2.1.2+), including polygon validity checks, spatial intersection, buffering, adjacency computation, and service-coverage queries. Unless otherwise specified, we use the default numerical tolerances and geometry predicates provided by GeoPandas and Shapely. Road-network graphs are constructed and analyzed with NetworkX (v3.5+), where nodes and edges are derived from the processed OSM road network obtained by OSMnx. Distance-related computations use the projected coordinate system of each study region, and metrics such as Service and Static Satisfaction are evaluated with the neighborhood-scale distance thresholds defined in Appendix~\ref{app:metric}.

For \model, each LiPUP cycle consists of one living stage followed by one planning stage.
In the living stage, we simulate 300 residents for one day under the current urban plan, with a simulation granularity of 5 minutes.
Resident profiles are automatically sampled and synthesized from aggregate census statistics of the selected region, providing demographic diversity without using individual-level personal records.
We build the living-stage simulator on AgentSociety~\cite{zhang2025parallelized} using its default settings, which provides a mature LLM-driven generative-agent backbone for resident profiles, daily activity simulation, and environment interaction.
The resident simulation is implemented with the DeepSeek-V4-Flash~\cite{deepseekv4flash} API.
In the planning stage, we use the GPT-5.4-mini~\cite{gpt54mini} API as the backbone LLM for the planner agent, with the temperature set to 0.6.
The same DeepSeek-V4-Flash API is also used to compute the Living Satisfaction metric in SimEval, with temperature set as 0, where resident agents evaluate their living experiences generated during simulation.
All experimental parameters are manually specified according to computational budget and evaluation stability, rather than being tuned through extensive hyperparameter search for maximizing performance.

In terms of token consumption, one LiPUP cycle requires approximately 60 million tokens for the living-stage simulation and approximately 1.25 million tokens for the planning stage.
Due to the high computational cost of resident simulation and its dependence on the generated plan, we do not repeat living-stage simulations for statistical aggregation. Instead, we run the planning stage of \model\ five times as stochastic candidate generation, select the plan with the highest Static Overall score, and evaluate only this selected plan with a single living-stage simulation. Therefore, the reported dynamic metrics and experience bank are obtained from this single simulation run of the selected plan, rather than from mean, maximum, or error-bar statistics over repeated living-stage simulations.

For baseline methods, we ensure a fair comparison by using the same GPT-5.4-mini API for all LLM-based planning methods and applying the same five-run selection protocol.
For optimization- and training-based baselines, including GA and DRL-GNN, we use the Static Overall score as the optimization objective, computed as the average of the three static metrics: Service, Ecology, and Static Satisfaction.
For DRL-GNN, all training and inference experiments are conducted on NVIDIA A40 GPUs.
Other baselines are evaluated under the same planning constraints and metric computation protocol as \model.

\section{Metric Details}
\label{app:metric}
We report three static automatic metrics and two SimEval metrics.
All metrics are computed for a plan $P: \mathcal{A}\rightarrow\mathcal{U}$ as defined in \secref{sec:pre}, where $\mathcal{A}=\{a_1,\ldots,a_{N_a}\}$ is the set of areas and $P(a_i)$ is the land-use type of area $a_i$.
Let $\mathcal{R}=\{r_1,\ldots,r_{N_r}\}$ denote the set of resident agents, and let $h_m$ be the home location of resident $r_m$.
Service and Static Satisfaction use road-network distance, while Ecology uses straight-line Euclidean distance.
For a land-use type $u\in\mathcal{U}$, we define the road-network distance from resident $r_m$ to the nearest area of type $u$ under plan $P$ as
$d_P^{\mathrm{road}}(r_m,u)=\min_{a_i\in\mathcal{A}: P(a_i)=u} \operatorname{dist}_{\mathrm{road}}(h_m,a_i)$,
where $\operatorname{dist}_{\mathrm{road}}(h_m,a_i)$ denotes the shortest-path distance in meters along the road network from $h_m$ to area $a_i$.
If no area of type $u$ exists in $P$, we set $d_P^{\mathrm{road}}(r_m,u)=+\infty$.

\noindent\textbf{Service.}
Following the static accessibility metric used in PUP, Service~\cite{zheng2023spatial} measures whether residents can conveniently access a predefined set of essential service land uses.
Let $\mathcal{U}_{\mathrm{svc}}\subseteq\mathcal{U}$ denote the set of basic service types, including education, healthcare, workplace, shopping, and entertainment facilities.
We compute
$\mathrm{Service}(P)=
\frac{1}{N_r |\mathcal{U}_{\mathrm{svc}}|}
\sum_{m=1}^{N_r}\sum_{u\in\mathcal{U}_{\mathrm{svc}}}
\mathbf{1}\left[d_P^{\mathrm{road}}(r_m,u)\leq 500\right]$,
where $\mathbf{1}[\cdot]$ is the indicator function.
Thus, the score is the average proportion of essential service types reachable within 500 meters from residents' homes along the road network.

\noindent\textbf{Ecology.}
Ecology measures the coverage of residents by parks and open spaces~\cite{zheng2023spatial} using straight-line Euclidean distance.
Let $\mathcal{U}_{\mathrm{eco}}\subseteq\mathcal{U}$ denote ecological land-use types, \ie parks.
We define the ecological service range under plan $P$ as
$\mathrm{ESR}(P)=
\bigcup_{a_i\in\mathcal{A}:\ P(a_i)\in\mathcal{U}_{\mathrm{eco}}}
\operatorname{Buffer}_{\mathrm{euc}}(a_i,300)$,
where $\operatorname{Buffer}_{\mathrm{euc}}(a_i,300)$ is the 300-meter Euclidean buffer around area $a_i$.
The Ecology metric is
$\mathrm{Ecology}(P)=\frac{1}{N_r}\sum_{m=1}^{N_r}
\mathbf{1}\left[h_m\in\mathrm{ESR}(P)\right]$.

\noindent\textbf{Static Satisfaction.}
Following the need-aware satisfaction metric in PUP, Static Satisfaction~\cite{zhou2024large} measures whether a plan satisfies residents' stated land-use preferences, without running the living simulation.
For each resident $r_m$, let $\mathcal{J}_m\subseteq\mathcal{U}$ be the set of preferred or needed land-use types collected before planning, according to the resident's profile.
The static satisfaction of resident $r_m$ is
$S_m^{\mathrm{static}}(P)=
\frac{1}{|\mathcal{J}_m|}
\sum_{u\in\mathcal{J}_m}
\mathbf{1}\left[d_P^{\mathrm{road}}(r_m,u)\leq 500\right]$,
and the overall metric is
$\mathrm{StaticSat}(P)=\frac{1}{N_r}\sum_{m=1}^{N_r}S_m^{\mathrm{static}}(P)$.

\noindent\textbf{Travel Distance.}
Travel Distance is the objective mobility-efficiency metric in SimEval.
After running the living simulator $g(\cdot)$ under plan $P$, each resident $r_m$ produces a sequence of mobility events $\mathcal{T}_m=\{\tau_{m,1},\ldots,\tau_{m,Q_m}\}$, where $Q_m$ is the number of movements made by resident $r_m$ and $\delta(\tau_{m,q})$ is the travel distance of movement $\tau_{m,q}$ in meters.
We compute the average distance over all realized resident movements:
$\mathrm{TravelDist}(P)=
\frac{1}{\sum_{m=1}^{N_r}Q_m}
\sum_{m=1}^{N_r}\sum_{q=1}^{Q_m}
\delta(\tau_{m,q})$.
A lower value indicates a lower daily mobility burden.

\noindent\textbf{Living Satisfaction.}
Living Satisfaction is the subjective SimEval metric.
After simulation, each resident $r_m$ receives a score $s_m\in\{1,2,3,4,5\}$ from the LLM-as-a-Judge protocol in \appref{app:prompt_eval}, based on the resident's simulated daily experience.
We compute $\mathrm{LivingSat}(P)=\frac{1}{N_r}\sum_{m=1}^{N_r}s_m$.

\section{Human Evaluation}
\label{app:humaneval}
In this section, we aim to evaluate the effectiveness of DeepSeek-V4-Flash in assessing the Living Satisfaction metric from residents' reflections and lived experience. 
We randomly sample 50 residents from Daxing, Beijing, and conduct a human verification study on the quality of reasoning outputs produced by the LLM evaluator. 
In addition, human annotators independently assess the same Living Satisfaction metric, enabling a direct comparison between LLM-based and human evaluations.
The full prompt used for LLM-based evaluation is provided in \appref{app:prompt_eval}.
Notably, human evaluators were given the same evaluation criteria and scoring instructions to ensure consistency between human and LLM assessments.

For the human evaluation, we invited three internal researchers who were familiar with the project and had relevant research background as evaluators.
Each evaluator independently annotated the same set of 50 residents, after which the annotations were aggregated and discussed to improve reliability. 
The evaluators participated voluntarily without additional payment. 
To comply with the double-blind review policy, we do not disclose their identities, specific roles, demographic information, or country of residence. 
Since the evaluators were internal project members and the study only involved annotation of already curated research materials, their participation and the intended use of the data were fully explained and agreed upon before the annotation.
Moreover, no separate ethics review board approval was sought, because this human evaluation was an internal expert annotation study rather than a new human-subject data collection procedure. It did not recruit external participants, interact with residents, or collect additional personal information.

Specifically, we report three alignment measures: (1) the agreement rate between LLM and human judgments; (2) Kendall's Tau correlation; and (3) Spearman's rank correlation between LLM-assigned and human-assigned satisfaction scores across the sampled residents. \tabref{table:human} summarizes the alignment between LLM and human evaluators in assessing Living Satisfaction.

\begin{table}[!h]
\centering
\caption{Alignment performance (\%).}
\vspace{-10pt}
\resizebox{0.9\linewidth}{!}{
\begin{tabular}{c|c|c}
\toprule
\textbf{Agree. Rate} & \textbf{Kendall Tau} & \textbf{Spearman Coeff.}\\
\hline
86.00 & 73.49 & 75.60\\
\bottomrule
\end{tabular}}
\vspace{-10pt}
\label{table:human}
\end{table}

\section{Skill Details}
\label{app:skill}
In \model{}, a skill is a reusable planning unit with two parts: an \emph{operating protocol} and a \emph{tool set}. The harness decides when a skill is called; the skill defines what it does once called. Compared with raw tools, skills give the agent a simpler interface and more stable reuse across planning steps.

\begin{table*}[t]
\centering
\caption{Tool-level specification of the multimodal skill library. Each tool is listed with its associated skill and intended function.}
\label{tab:skill_tool_map_full}
\small
\renewcommand{\arraystretch}{1.15}
\setlength{\tabcolsep}{4pt}

\begin{tabularx}{\textwidth}{
>{\raggedright\arraybackslash}p{2.25cm}
>{\raggedright\arraybackslash}p{4.8cm}
>{\raggedright\arraybackslash}X
}
\toprule
\textbf{Skill} & \textbf{Tool} & \textbf{Function} \\
\midrule

\multirow{6}{*}{\parbox[c][3.9cm][c]{2.25cm}{\raggedright Experience\\Diagnosis}}
& \texttt{retrieve\_hotspot\_areas} & Retrieves high-salience areas from aggregated resident activity and complaint intensity. \\
\cline{2-3}
& \texttt{profile\_area\_urgency} & Estimates renewal urgency for queried areas from recurring area-level dissatisfaction signals. \\
\cline{2-3}
& \texttt{profile\_neighbor\_externalities} & Identifies dissatisfaction caused by surrounding land uses and spillover effects from nearby areas. \\
\cline{2-3}
& \texttt{filter\_areas\_by\_land\_use} & Selects areas with specified land-use types to support targeted experiential analysis. \\
\cline{2-3}
& \texttt{summarize\_regional\_experience} & Aggregates resident feedback within a region into a compact summary of recurring local issues. \\
\cline{2-3}
& \texttt{summarize\_transition\_frictions} & Summarizes mobility-related frictions such as inconvenient or inefficient inter-area travel experiences. \\
\midrule

\multirow{6}{*}{\parbox[c][3.9cm][c]{2.25cm}{\raggedright Visual\\Contextualization}}
& \texttt{render\_global\_plan} & Renders the full urban plan for global inspection of functional distribution and large-scale layout. \\
\cline{2-3}
& \texttt{render\_local\_zoom} & Produces zoomed-in visualizations of selected regions for local land-use inspection. \\
\cline{2-3}
& \texttt{highlight\_candidate\_areas} & Overlays queried areas on the map to visually anchor candidate targets or evidence locations. \\
\cline{2-3}
& \texttt{overlay\_revision\_targets} & Marks proposed revision targets together with their surrounding context for comparative inspection. \\
\cline{2-3}
& \texttt{caption\_spatial\_pattern} & Generates textual descriptions of visible spatial structures, adjacencies, and layout patterns. \\
\cline{2-3}
& \texttt{compare\_plan\_versions} & Visually compares alternative plan states or pre/post-revision layouts. \\
\midrule

\multirow{6}{*}{\parbox[c][3.9cm][c]{2.25cm}{\raggedright Geospatial\\Grounding}}
& \texttt{query\_area\_details} & Returns structured attributes of queried areas, including land use, geometry, and local context. \\
\cline{2-3}
& \texttt{query\_areas\_in\_bbox} & Retrieves all areas located within a specified geographic bounding box. \\
\cline{2-3}
& \texttt{query\_adjacent\_areas} & Returns neighboring or topologically adjacent areas relevant to the queried location. \\
\cline{2-3}
& \texttt{estimate\_inter\_area\_distance} & Computes straight-line or network distance between selected areas. \\
\cline{2-3}
& \texttt{check\_land\_use\_compatibility} & Evaluates whether candidate land-use assignments are spatially compatible with nearby functions. \\
\cline{2-3}
& \texttt{inspect\_service\_coverage} & Assesses local service accessibility and coverage conditions around queried residential areas. \\
\midrule

\multirow{6}{*}{\parbox[c][3.9cm][c]{2.25cm}{\raggedright Revision\\Verification}}
& \texttt{apply\_candidate\_revision} & Applies tentative land-use changes to a working plan state for downstream evaluation. \\
\cline{2-3}
& \texttt{validate\_planning\_constraints} & Checks hard planning constraints, including immutability rules and modification-budget limits. \\
\cline{2-3}
& \texttt{evaluate\_revision\_metrics} & Returns planning metrics for a candidate revision, including service, ecology, and resident-centered outcomes. \\
\cline{2-3}
& \texttt{audit\_cumulative\_revisions} & Lists all cumulative changes relative to the initial baseline for global consistency checking. \\
\cline{2-3}
& \texttt{rollback\_to\_snapshot} & Restores the working plan to a previously saved revision state when a candidate is rejected. \\
\cline{2-3}
& \texttt{compare\_revision\_snapshots} & Compares multiple tentative revision states to support evidence-based selection among alternatives. \\

\bottomrule
\end{tabularx}
\end{table*}

\tcbset{
    width=\linewidth, 
    colback=gray!10, 
    colframe=black, 
    coltitle=white, 
    fonttitle=\bfseries, 
    arc=3mm, 
    colbacktitle=gray!120, 
    boxsep=5pt, 
    boxrule=0.8pt, 
    left=0pt, 
    right=0pt, 
    breakable,
    enhanced jigsaw,
    fontupper=\ttfamily, 
}

\begin{tcolorbox}[title=Experience Diagnosis]
\small
\textbf{Skill summary.} Find problem areas, recurring local needs, neighbor spillovers, and mobility frictions from PGeB.

\textbf{Operating protocol.} Use hotspot, urgency, neighbor, or land-use filters to find relevant AOIs. For a target AOI or region, query evidence and summarize self-area needs, spillovers, and mobility frictions. If the scope is too broad, narrow it by AOI list or bbox. Return candidate AOIs, issues, and confidence. Do not propose land-use edits unless the evidence supports them.
\end{tcolorbox}

\begin{tcolorbox}[title=Visual Contextualization]
\small
\textbf{Skill summary.} Inspect global layout, local structure, target locations, and pre/post-revision map consistency.

\textbf{Operating protocol.} Render the global plan first to see the overall layout, then zoom into candidate regions or highlighted AOI IDs. Inspect both the initial and current modified map when revisions accumulate. Describe clustering, isolation, adjacency conflicts, service gaps, and whether candidate edits remain coherent. Return a visual grounding note tied to exact AOI IDs or bbox coordinates.
\end{tcolorbox}

\begin{tcolorbox}[title=Geospatial Grounding]
\small
\textbf{Skill summary.} Check AOI attributes, mutability, adjacency, distances, service coverage, and land-use compatibility.

\textbf{Operating protocol.} Query AOI details before editing to confirm land use, mutability, centroid, and local evidence. Use bbox, related-AOI, and distance tools to check nearby functions, mobility coupling, and accessibility. Use road distance for service access and straight-line distance for local adjacency or ecology checks. If visual and structured evidence conflict, use the structured result. Return a feasibility profile that marks candidate edits as allowed, redundant, risky, or supported.
\end{tcolorbox}

\begin{tcolorbox}[title=Revision Verification]
\small
\textbf{Skill summary.} Test tentative edits, check constraints and metrics, compare revision states, and roll back unsupported trials.

\textbf{Operating protocol.} Work only after candidate edits are supported by evidence and constraints. Apply tentative land-use changes in small batches with exact AOI IDs and target land-use codes. Check invalid inputs, no-change entries, cumulative modifications, map feedback, and Service/Ecology/Satisfaction metrics after each call. If a batch harms the plan or violates constraints, undo it by restoring the original land use or test a smaller alternative. Before finalizing, audit all cumulative modifications and return either an accepted update set with metric support or a rejection note.
\end{tcolorbox}

\section{Artifact Statement}
\label{app:artifact}

\noindent\textbf{Urban data.}
We use two types of public urban data. First, we collect geographic and land-use information from OpenStreetMap (OSM)\footnote{\url{https://www.openstreetmap.org/copyright}} for the selected study regions and preprocess it into area-level urban plans, spatial relations, and map-based inputs. OSM data are available under the Open Data Commons Open Database License (ODbL), which allows use and adaptation with attribution to OpenStreetMap contributors and share-alike requirements for derived databases. If released, our processed OSM-derived planning data will preserve the required attribution and license compatibility. These artifacts cover two community-level regions in Daxing, Beijing and Adlershof, Berlin, and document spatial units, land-use categories, adjacency, distance, and accessibility-related attributes for urban planning research.

Second, we use public aggregate demographic statistics to synthesize resident profiles. For Beijing, we refer to the publicly released \textit{Beijing Population Census Yearbook 2020}\footnote{\url{https://nj.tjj.beijing.gov.cn/tjnj/rkpc-2020/indexch.htm}}, which reports aggregate population structures such as gender and age, education, housing, economic activity, marriage and fertility, and township-level population statistics. We use these tables only to estimate population-level distributions for synthetic resident attributes, such as age group, gender, education level, occupation or employment status, household-related attributes, and residential background when applicable. We do not redistribute the original census tables. The generated profiles are synthetic simulation artifacts rather than records of real individuals. Both the OSM-derived planning data and census-based profile distributions are public, aggregate, and non-personal. They do not contain private user records, personally identifying information, or offensive natural-language content, so no additional anonymization or offensive-content filtering is required beyond restricting preprocessing to planning-relevant geographic attributes and aggregate demographic distributions.

\noindent\textbf{Simulation and baseline code.}
We use code artifacts for simulation, baseline comparison, preprocessing, planning, and evaluation. AgentSociety\footnote{\url{https://github.com/tsinghua-fib-lab/agentsociety/blob/main/LICENSE}} is used as the simulation backbone for generating resident daily activities and living experiences. Its public GitHub repository is released under Apache-2.0, except for the repository's commercial folder, and its intended use includes LLM-based agent simulation and research experimentation. This is compatible with our use for research simulation in participatory urban planning.

For DRL-GNN, we use the public DRL-urban-planning implementation\footnote{\url{https://github.com/tsinghua-fib-lab/DRL-urban-planning/blob/main/LICENSE}} released by the original authors under the MIT License. This repository provides reproduction code for spatial planning of urban communities via deep reinforcement learning, and we use it only as a research baseline implementation. Other baselines, including Initial, Random, GA, and PUP-MA, do not involve reusing additional released model checkpoints or external code artifacts in our experiments. They are either simple algorithmic baselines or our own implementations following the corresponding papers.

Our own code artifacts, including OSM preprocessing, profile synthesis, LiPUP-MA implementation, and metric calculation for both static and simulation-based evaluation, are intended for research on living-in-the-loop participatory urban planning. If released, they will be provided under a permissive research-compatible license, while preserving the attribution and usage requirements of the upstream code and data sources.

\section{Statement on the Use of AI Assistance}
All research ideas, experimental design, logical reasoning, and core writing were originally conceived and carried out by our authors. 
AI tools, specifically Codex for coding assistance, and GPT/DeepSeek for manuscript polishing, were used only as auxiliary aids. 
GPT Image-2 was employed to generate certain icons in \figref{fig:cup} and \figref{fig:main}, solely for visual enhancement and readability.
These icons carry no data-related information and involve no manipulation or fabrication of results. 
The same tool also produced example figures as layout references, but all final figures were manually drawn and composed by the authors. 
The overall work was primarily completed by our authors, with AI serving only a supportive role.

\section{Prompts}
\label{app:prompt}

\subsection{\model: Simulation Stage}
We report the simulation-stage prompt used by \model. Routine daily activity planning and action execution prompts come from the simulator backbone and are not repeated.

\begin{tcolorbox}[title=Urban-grounded Reflection]
\small
You are a resident agent. After one mobility-based activity under the current urban plan, turn that activity into planning evidence for PGeB. Do not write a generic diary.\\

Inputs:\\
Resident profile: \texttt{<RESIDENT\_PROFILE>}.\\
Completed activity: \texttt{<COMPLETED\_ACTIVITY>}.\\
Urban context: \texttt{<URBAN\_CONTEXT>}.\\
Past comments: \texttt{<PAST\_AREA\_COMMENTS>}.\\

Rules:\\
1. Ground every comment in the completed activity, your profile, and the urban context. Do not invent urban facts.\\
2. For area experience, comment on the AOI or its surrounding land-use pattern, such as service adequacy, convenience, comfort, environmental quality, or nearby conflicts.\\
3. For transition experience, judge whether the trip between origin and destination fits the activity. Pay attention to travel distance and time; service access beyond about 500 meters or 15 minutes can be treated as a burden, depending on resident tolerance and activity type.\\
4. Update past comments when the latest activity adds new evidence. Return only AOIs or transitions that provide useful planning evidence. If there is no planning-relevant experience, return empty lists.\\
5. Keep each comment short, specific, and in English. Output JSON only.\\

Answer format in JSON. Area items: \texttt{aoi\_id}, \texttt{scope}, \texttt{sentiment}, \texttt{comment}, \texttt{evidence}. Transition items: \texttt{origin\_aoi}, \texttt{destination\_aoi}, \texttt{purpose}, \texttt{distance\_m}, \texttt{time\_min}, \texttt{sentiment}, \texttt{comment}.\\
\texttt{\{"area\_experience":[\{...\}],}\\
\texttt{"transition\_experience":[\{...\}]\}}
\end{tcolorbox}

\subsection{\model: Planning Stage}

\subsubsection{General Prompts}

\begin{tcolorbox}[title=Skill Selection]
\small
You are an autonomous urban renewal planning agent. Complete the current step by selecting the next skill, or finish the step when the required output is ready. The workflow is a repeated locate-ground-execute loop. Each loop must improve the land-use plan under these constraints: RS and OS areas are immutable; only BS, RC, PK, OF, SC, HS, and ID may be converted to one another; final modified AOIs must not exceed 30\% of all modifiable AOIs. Optimize Service, Ecology, and Satisfaction together, with special attention to the weakest metric.\\

Inputs:\\
History: \texttt{<SKILL\_LEVEL\_HISTORY\_CONTEXT>}.\\
Step prompt: \texttt{<STAGE\_WISE\_PROMPT>}.\\
Skills: \texttt{<AVAILABLE\_SKILLS>}.\\

Decision rules:\\
1. Identify what information or action is still missing.\\
2. Select exactly one skill if another skill result is needed. Prefer the skill that most directly closes the gap.\\
3. End the step only when the current step prompt can be satisfied from accumulated evidence. Do not end with unsupported assumptions.\\
4. Avoid repeated skill calls that do not reduce uncertainty. If a previous skill result is insufficient, state the gap.\\

Answer Format in JSON:\\
\texttt{\{"thought":"brief rationale", "action":"select\_skill", "skill\_name":"Experience Diagnosis|Visual Contextualization|Geospatial Grounding|Revision Verification", "instruction":"what this skill should accomplish next"\}}\\

or\\

\texttt{\{"thought":"brief rationale", "action":"finish\_step", "step\_result":\{...\}\}}
\end{tcolorbox}

\begin{tcolorbox}[title=Skill Execution]
\small
You are executing one selected planning skill inside the current harness step. Follow the selected skill's operating protocol, use the available tools in ReAct style, and return a skill result only when the skill has enough evidence or feedback for the step controller.\\

Inputs:\\
History: \texttt{<SKILL\_LEVEL\_HISTORY\_CONTEXT>}.\\
Step prompt: \texttt{<STAGE\_WISE\_PROMPT>}.\\
Protocol: \texttt{<SKILL\_OPERATING\_PROTOCOL>}.\\
Tool history: \texttt{<TOOL\_LEVEL\_HISTORY\_CONTEXT>}.\\
Tools: \texttt{<AVAILABLE\_TOOLS>}.\\

Tool-use rules:\\
1. Use exact AOI IDs returned by tools. Verify mutability before proposing or applying edits.\\
2. Use visual tools with query tools for grounding; use zoomed views for local regions and current-map renderings after temporary edits.\\
3. For spatial queries, use absolute coordinates in meters. Do not pass normalized bbox hints to tools that require x1, y1, x2, y2.\\
4. Apply tentative plan changes only after evidence and constraints are checked. Each temporary update is cumulative until explicitly undone.\\
5. If a tool returns an error, invalid input, illegal edit, or no-change record, correct the issue before continuing.\\
6. Keep the skill result compact: preserve evidence, AOI IDs, metric feedback, accepted or rejected edits, and unresolved gaps; omit unnecessary intermediate reasoning.\\

Answer Format in JSON for a tool call:\\
\texttt{\{"thought":"why this tool is needed", "action":"call\_tool", "tool":"TOOL\_NAME", "args":\{...\}\}}\\

Answer Format in JSON for returning from the skill:\\
\texttt{\{"thought":"why the skill can stop", "action":"finish\_skill", "skill\_result":\{...\}\}}
\end{tcolorbox}

\subsubsection{Step-wise Prompts}

\begin{tcolorbox}[title=Revision Target Locating]
\small
Current step: Revision Target Locating. Find planning-worthy AOIs or small neighboring regions for this loop. Use resident experience signals, mobility hotspots, current land-use distribution, and visual layout evidence to update the global target list and select priority targets. Prefer targets with recurring complaints, neighbor spillovers, high activity, clear relevance to Service, Ecology, and Satisfaction, and plausible modifiability. Do not propose final land-use changes in this step.\\

Required step output:\\
\texttt{\{"global\_target\_list":[...], "priority\_targets":[\{"target\_id":"...", "aoi\_ids":[...], "reason":"...", "evidence\_summary":"...", "remaining\_uncertainty":"..."\}], "next\_step\_focus":"..."\}}
\end{tcolorbox}

\begin{tcolorbox}[title=Multimodal Evidence Grounding]
\small
Current step: Multimodal Evidence Grounding. For the priority targets selected in Step 1, collect and organize evidence only. Combine resident evidence, visual context, AOI details, nearby AOIs, distance and accessibility relations, and land-use constraints to clarify needs, spatial opportunities, risks, and unresolved uncertainty. Do not propose concrete land-use edits or test revisions in this step; leave trial decisions to Reflective Execution.\\

Required step output:\\
\texttt{\{"grounded\_targets":[\{"target\_id":"...", "aoi\_ids":[...], "resident\_needs":"...", "visual\_context":"...", "geospatial\_constraints":"...", "feasibility\_notes":"supported|risky|unsupported with reasons", "remaining\_uncertainty":"..."\}], "evidence\_priority\_for\_execution":[...]\}}
\end{tcolorbox}

\begin{tcolorbox}[title=Reflective Execution]
\small
Current step: Reflective Execution. Use the grounded evidence from Step 2 to propose concrete land-use edits, test them through tentative updates, inspect map feedback, evaluate metrics, and audit cumulative modifications. Iterate with small batches, verify hard constraints and Service, Ecology, and Satisfaction effects, then accept, revise, or roll back the batch. If a trial is invalid, harmful, redundant, or spatially incoherent, reject it or undo it before trying another alternative. Finish planning only when the plan is coherent, the modification budget is respected, and further trials are unlikely to improve the objectives.\\

Required step output for continuing another loop:\\
\texttt{\{"accepted\_edits":\{"AOI\_ID":"LANDUSE"\}, "rejected\_trials":\{...\}, "trial\_rationale":"...", "metric\_feedback":"...", "remaining\_targets":[...], "continue\_planning":true\}}\\

Required final output when planning is complete:\\
\texttt{\{"final\_plan":\{"AOI\_ID":"LANDUSE", ...\}, "accepted\_edits":\{...\}, "rejected\_trials":\{...\}, "metric\_feedback":"...", "constraint\_check":"...", "continue\_planning":false\}}
\end{tcolorbox}

\subsection{SimEval}
\label{app:prompt_eval}

\begin{tcolorbox}[title=Living Satisfaction Judging]
\small
You are an urban life evaluator. Based on one resident's simulated daily experience under the current plan, assign an overall living satisfaction score for that resident.\\

Inputs:\\
Resident profile: \texttt{<RESIDENT\_PROFILE>}.\\
Simulated trajectories: \texttt{<SIMULATED\_TRAJECTORIES>}.\\
Urban grounded reflections: \texttt{<URBAN\_GROUNDED\_REFLECTIONS>}.\\
Current plan context: \texttt{<CURRENT\_PLAN\_CONTEXT>}.\\
Previous simulation reflections: \texttt{<PREVIOUS\_SIMULATION\_REFLECTIONS>}.\\

Rules:\\
1. Score only from the simulated trajectories and reflections under the current plan. Do not rely on static preferences alone.\\
2. Consider daily mobility burden, service access, land-use match to activities, environmental comfort, nearby conflicts, and repeated satisfaction or dissatisfaction in reflections.\\
3. If previous simulation reflections are provided, use them only as a baseline. Reward clearly improved current experiences and penalize unresolved or worsened problems when supported by current evidence.\\
4. The score must be an integer from 1 to 5, where 1 means highly dissatisfied, 3 means mixed or acceptable with clear issues, and 5 means highly satisfied. A resident with major unresolved accessibility, service, or environmental problems should not receive 4 or 5.\\
5. Think internally and output JSON only.\\

Answer format in JSON:\\
\texttt{\{"score":1, "reason":"short evidence-grounded reason"\}}
\end{tcolorbox}

\end{document}